\pdfoutput=1

\documentclass[11pt]{article}

\usepackage[final]{acl}

\usepackage{times}
\usepackage{latexsym}
\usepackage{booktabs} 
\usepackage{amsmath} 
\usepackage{booktabs}
\usepackage{xcolor}
\usepackage[table,xcdraw]{xcolor}

\usepackage{arydshln} 

\usepackage{kotex}
\usepackage{multirow}

\usepackage{booktabs}
\usepackage{caption}
\usepackage{subcaption}
\usepackage{amssymb}
\usepackage{tikz}

\usepackage[T1]{fontenc}

\usepackage[utf8]{inputenc}

\usepackage{microtype}

\usepackage{inconsolata}

\usepackage{graphicx}

\title{Semantic Hardness Is Not Visual Hardness: Sign-Aware Hard Negative Mining for Sign Language Retrieval}

\newcommand\CorrespondingAuthorMark{\footnotemark[\value{footnote}]}

\author{
Junmyeong Lee$^1$\hspace{2.5mm}        
Chan Hur$^4$\hspace{2.5mm}                                                                   ChangSu Choi$^2$\hspace{2.5mm}                                                               Sukmin Cho$^1$\hspace{2.5mm} \\                                                              \textbf{Fitsum Gaim}$^1$\hspace{2.5mm}                                                       \textbf{Eui Jun Hwang}$^1$\hspace{2.5mm}                                                     \textbf{Hoyun Song}$^3$\thanks{Corresponding Author}\hspace{2.5mm}  
\textbf{KyungTae Lim}$^{2,3}$\protect\CorrespondingAuthorMark \hspace{2.5mm}\\[0.3mm] 
School of Computing$^1$\hspace{2.5mm}                                                 
Graduate School of Culture Technology$^2$\hspace{2.5mm}\\
InnoCORE PRISM-AI Center$^3$\hspace{2.5mm}   
Korea Advanced Institute of Science and Technology$^1$$^,$$^2$$^,$$^3$\hspace{2.5mm}\\
ETRI Medical Informatics Laboratory$^4$\hspace{2.5mm}\\[0.3mm]
\texttt{david516@kaist.ac.kr}\hspace{2.5mm}\texttt{chanhur@etri.re.kr}\\
\texttt{\{choics2623,nelllpic,fitsum.gaim,ehwa20,hysong,ktlim\}@kaist.ac.kr}\\
}

\begin{document}
\maketitle
\begin{abstract} 

Sign Language Retrieval (SLRet) enables efficient access to sign language content but remains fragile in fine-grained scenarios where visually similar signs must be distinguished. We show that this limitation does not stem from model capacity, but from ineffective hard negative supervision. Specifically, we formulate fine-grained retrieval failures as a negative distribution mismatch: semantically distinct yet visually confusable signs are rarely treated as hard negatives, while existing text-based mining strategies fail to capture such visual ambiguity. To address this issue, we propose Sign-Aware Hard Negative Mining (SAN), which constructs hard negatives based on visual confusability in the sign embedding space rather than linguistic similarity. Experiments on PHOENIX-2014T demonstrate that SAN substantially improves fine-grained retrieval performance while preserving coarse-grained accuracy, highlighting the importance of aligning negative supervision with visual ambiguity in sign language retrieval. Code is available at Github repository.\footnote{\url{https://github.com/joonmy/SAN.git}}.
\end{abstract}

\section{Introduction}

Sign languages are the primary means of communication for the Deaf community, expressed through hand, body, and facial movements. The unique grammar and visual complexity of sign languages often create communication barriers between signers and non-signers. To bridge this gap, prior research has focused on sign language understanding, particularly sign language recognition~\cite{DBLP:conf/iccv/slr_ref_1,DBLP:conf/cvpr/slr_ref_3,DBLP:conf/cvpr/slr_ref_2} and translation~\cite{DBLP:conf/cvpr/slt_ref_2, DBLP:conf/iccv/slt_ref3, DBLP:conf/cvpr/slt_ref_7}. However, the scarcity of sign language data leads to high error rates across both tasks~\cite{DBLP:conf/cvpr/cico}.

Recently, Sign Language Retrieval (SLRet)\ \cite{DBLP:conf/cvpr/spot-align,DBLP:conf/cvpr/cico, DBLP:conf/eccv/UPRet} has emerged as a promising task. SLRet aims to retrieve relevant sign language videos or texts from a database given a query. It enables efficient access to sign language content and facilitates the use of unannotated sign videos, helping mitigate the data scarcity of sign language resources~\cite{DBLP:conf/cvpr/spot-align}.

However, we identify a core bottleneck that current SLRet models fail to address. Since sign languages construct meaning within a restricted visual space, even subtle differences in hand shape, position, or trajectory can alter meaning—a phenomenon known as \textit{sign confusability}~\cite{DBLP:conf/eccv/slr_ref_5, DBLP:conf/cvpr/slr_ref_2}. While existing models remain stable in coarse-grained retrieval, their performance degrades sharply in fine-grained retrieval scenarios where distinguishing these subtle distinctions is required. We argue that this degradation stems not from a lack of model capacity, but rather from training supervision that does not sufficiently expose the model to such cases.

\begin{figure}[t] 
    \centering
    \includegraphics[width=1\columnwidth]{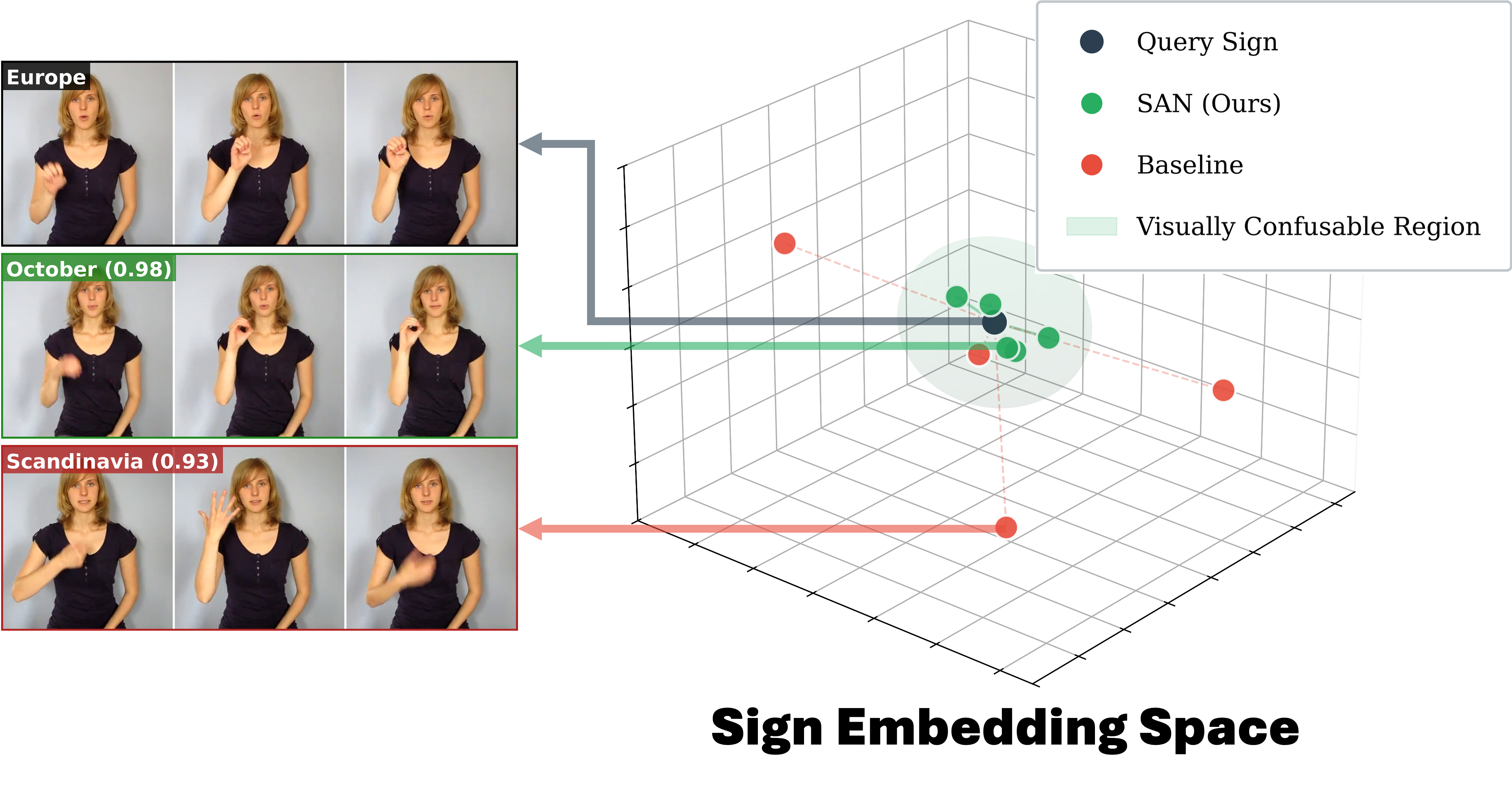}
    \caption{\label{fig:motivation} Illustration of fine-grained ambiguity in sign language retrieval.
    Semantically distinct words often correspond to visually similar signs, forming true hard negatives. SAN effectively targets these visually confusable instances that text-based mining methods frequently fail to address.}
    \vspace{-0.2in}
\end{figure}

We formulate the cause of this failure as a \textit{negative distribution mismatch} in contrastive learning, which is exacerbated in sign language retrieval due to frequent sign confusability. Importantly, this mismatch manifests in two forms. First, sign language datasets remain primarily coarse-grained~\cite{DBLP:conf/bmvc/scarcity_3,DBLP:conf/cvpr/scarcity_4}, lacking sufficient naturally occurring hard negatives to support fine-grained feature learning. Second, while previous studies have attempted to alleviate this limitation by generating hard negative captions through text-based perturbations~\cite{DBLP:conf/iccv/verb_in_action, DBLP:conf/accv/beyond, DBLP:conf/cvpr/ranking_clip}, such approaches may still provide insufficient guidance for resolving sign confusability, as they rely on linguistic semantics rather than visual similarity.

Figure~\ref{fig:motivation} intuitively illustrates this discrepancy. For a query word \textit{Europe}, traditional text-based approaches select semantically related words like \textit{Scandinavia} as hard negatives. However, because their corresponding signs are visually distinct, they serve merely as easy negatives for the model. By contrast, words like \textit{October}, which are semantically unrelated but visually similar in their sign form, act as true hard negatives that the model struggles to distinguish. Yet such visually confusable samples are not covered by text-based negative mining approaches. Consequently, existing approaches fail to expose the model to learn the fine-grained differences located in the \textit{Visually Confusable Region}.

Motivated by this observation, we propose \textbf{Sign-Aware Hard Negative Mining (SAN)}, which redefines the criteria for hard negative mining from linguistic similarity to visual confusability. SAN operates by (1) extracting high-confidence sign–word correspondences, (2) identifying signs that cluster closely in the sign language embedding space yet are semantically distinct, and (3) constructing hard negative captions using the words corresponding to these visually confusable signs, thereby correcting biases in the supervision signals.

We evaluate SAN on the PHOENIX-2014T dataset~\cite{DBLP:conf/cvpr/slt_ref1} and present three key findings. First, existing SLRet models exhibit substantial performance degradation in fine-grained retrieval scenarios involving visually confusable queries. Second, text-based negative mining fails to adequately address this problem. Third, SAN significantly improves fine-grained retrieval performance while preserving coarse-grained accuracy. These findings suggest that fine-grained discrimination failures in sign language retrieval stem from a supervision mismatch, and that aligning negatives with visual ambiguity is key to resolving it.

Our contributions are summarized as follows: 
\begin{itemize} 
    \item We formulate fine-grained retrieval failures in sign language retrieval as a \textit{negative distribution mismatch} problem.
    \item We propose Sign-Aware Hard Negative Mining (SAN), which aligns hard negatives with visual confusability in sign language.
    \item We demonstrate that SAN substantially improves fine-grained retrieval while maintaining stable coarse-grained performance.
\end{itemize}



\section{Related Work}
\subsection{Sign Language Retrieval}

Early research on sign language understanding primarily focused on Sign Language Recognition (SLR)~\cite{DBLP:conf/cvpr/slr_ref_2, DBLP:conf/aaai/slr_ref_4, DBLP:conf/icassp/slr_ref_cslr1, DBLP:journals/pr/slr_ref_cslr5} and Sign Language Translation (SLT)~\cite{DBLP:conf/iclr/slt_ref_glossbased1, DBLP:conf/cvpr/slt_ref_7, DBLP:conf/iccv/slt_ref3, DBLP:conf/naacl/slt_ref6}. SLR aims to predict glosses from sign language videos, whereas SLT focuses on generating natural language sentences.

More recently, Sign Language Retrieval (SLRet) has emerged as an important direction in sign language research. SLRet aims to retrieve semantically relevant videos or captions from sign language datasets given a query. This capability is particularly important given the rapid growth of online sign language content as it enables efficient access to sign resources~\cite{DBLP:conf/cvpr/spot-align, DBLP:conf/cvpr/cico}. 
SPOT-ALIGN~\cite{DBLP:conf/cvpr/spot-align} pioneered the SLRet task by introducing global alignment between sign videos and textual captions. Subsequently, CiCo~\cite{DBLP:conf/cvpr/cico} advanced this paradigm by redefining SLRet as a cross-lingual retrieval problem and introducing cross-lingual contrastive learning (CLCL) to achieve finer alignment between sign units and textual tokens. SEDS~\cite{DBLP:conf/mm/SEDS} further enhanced this approach by incorporating an additional keypoint modality for joint intra- and inter-modal learning. UPRet~\cite{DBLP:conf/eccv/UPRet} addressed sign ambiguity by modeling both modalities as Gaussian distributions and aligning them via optimal transport.

However, a critical limitation persists: these methods predominantly optimize for coarse-grained retrieval, where in-batch negatives are relatively easy to distinguish. As a result, current models remain limited in fine-grained retrieval settings, where subtle visual distinctions between signs are crucial. Despite its importance, this aspect has been largely underexplored, and we therefore focus on this challenge.

\subsection{Hard Negatives in Vision--Language Models}
Hard negatives provide informative supervision for contrastive learning by exposing models to hard-to-distinguish examples. In vision--language research, a common strategy to achieve this is to construct \emph{hard negative captions}. Most prior works generate such negatives by modifying specific words in the original captions within the \emph{textual domain}. To this end, standard approaches utilize rule-based heuristics (e.g., word swapping)~\cite{DBLP:negclip, DBLP:teaching_clip_count_to_ten, DBLP:conf/accv/beyond} or leverage pretrained language models to synthesize counterfactual captions~\cite{DBLP:conf/cvpr/teaching_structured_clip, DBLP:conf/iccv/verb_in_action, DBLP:conf/cvpr/ranking_clip, DBLP:conf/nips/tripletclip}. These methods implicitly assume that semantically difficult negatives (e.g., changing `man' to `boy') act as effective training signals for fine-grained visual discrimination. 

While text-driven negative mining has proven effective in general vision--language domains, it is often suboptimal for sign language. In sign language, semantically related words are often expressed through entirely different motions, whereas semantically unrelated words can share high visual similarity~\cite{DBLP:conf/cvpr/slr_ref_2}. Furthermore, because sign language conveys meaning through a constrained set of gestures, visually similar signs occur frequently~\cite{DBLP:conf/eccv/slr_ref_5}. As a result, text-based mining methods—operating solely in the textual semantic space—tend to produce \emph{visually easy} negatives, failing to provide informative supervision for fine-grained visual discrimination.

To bridge this gap, we propose a sign-aware hard negative mining method that explicitly accounts for visual similarity of signs, identifying truly confusable negatives beyond textual semantics.




\section{Methodology}
\begin{figure*}

    \centering{\includegraphics[width=\textwidth]{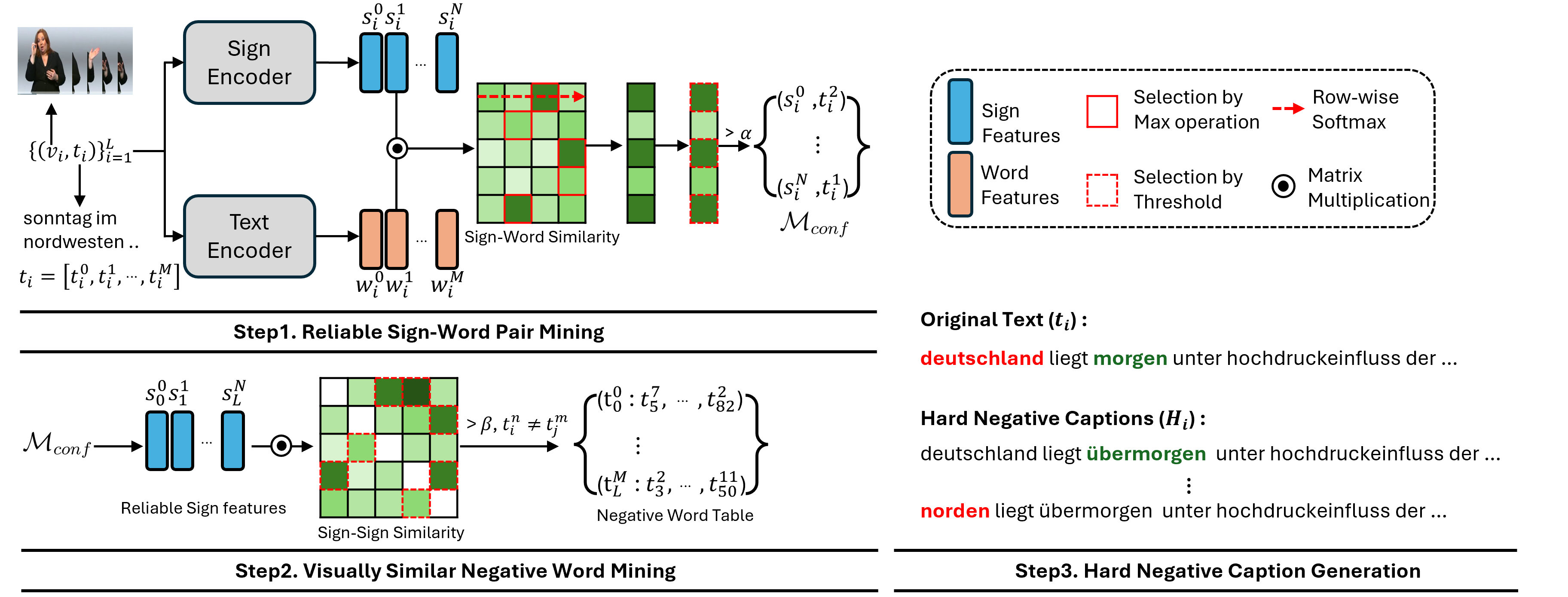}}

    \caption{
    Overview of the Sign-Aware Hard Negative Mining (SAN) framework. SAN consists of three steps: (1) Reliable Sign–Word Pair Mining to extract high-confidence alignments; (2) Visually Similar Negative Word Mining to identify visually confusable but semantically distinct words; and (3) Hard Negative Caption Generation to construct challenging training samples via keyword substitution.
    }
\end{figure*}

In this section, we first introduce the preliminaries of sign language retrieval task in Section~\ref{Sec3.1}. 
We then present our proposed Sign-Aware Hard Negative Mining framework in Section~\ref{Sec3.2}.

\subsection{Preliminary}
\label{Sec3.1}

Sign language retrieval aims to learn a shared embedding space that aligns sign videos $\mathcal{V}$ and textual descriptions $\mathcal{T}$ for cross-modal matching. We consider two standard settings: text-to-video (T2V) and video-to-text (V2T) retrieval, which retrieve the most relevant sign video $v \in \mathcal{V}$ for a given text query and the corresponding text $t \in \mathcal{T}$ for a sign video query, respectively.


Following prior work, we adopt the cross-lingual contrastive learning (CLCL) objective~\cite{DBLP:conf/cvpr/cico} to capture fine-grained alignments between sign units and word tokens. Given a mini-batch of $B$ video--text pairs $\{(v_i, t_i)\}_{i=1}^B$, we encode sign videos and texts using a visual encoder $F$ and a text encoder $G$, producing sign-level and word-level feature sequences
$v'_i = [s_i^0, s_i^1, \ldots, s_i^N]$ and $t'_i = [w_i^0, w_i^1, \ldots, w_i^M]$, where $N$ and $M$ denote the number of sign clips and words in the text, respectively. We then compute the sign--word similarity matrix and apply attention-weighted aggregation to obtain video--text similarity scores $S_{V2T}$ and $S_{T2V}$ for all video--text pairs in the mini-batch.

We optimize the retrieval model using the InfoNCE loss~\cite{DBLP:journals/jmlr/infonce}:
\begin{equation}
\small
\mathcal{L}_{\textit{V2T}} = -\frac{1}{B} \sum_{i=1}^{B} 
\log \frac{\exp(S_{V2T}(v_i, t_i)/\tau)}{\sum_{j=1}^{B} \exp(S_{V2T}(v_i, t_j)/\tau)},
\end{equation}

\begin{equation}
\small
\mathcal{L}_{\textit{T2V}} = -\frac{1}{B} \sum_{i=1}^{B} 
\log \frac{\exp(S_{T2V}(v_i,t_i)/\tau)}{\sum_{j=1}^{B} \exp(S_{T2V}(v_j, t_i)/\tau)},
\end{equation}

\begin{equation}
\small
\mathcal{L_\textit{coarse}} = \mathcal{L}_{\textit{V2T}} + \mathcal{L}_{\textit{T2V}},
\end{equation}
where $\tau$ is a learnable temperature parameter.

\subsection{Sign-Aware Hard Negative Mining}
\label{Sec3.2}
To address the mismatch between the negative supervision signals and the visual ambiguity inherent in sign language, we propose \emph{Sign-Aware Hard Negative Mining (SAN)}. From a contrastive learning perspective, SAN re-aligns the negative sample distribution with visually confusable regions of the sign space, which are underrepresented by standard mini-batch negatives. SAN consists of three components: (1) reliable sign--word pair mining, (2) visually similar negative word mining, and (3) hard negative caption generation.

\paragraph{Reliable Sign--Word Pair Mining}

To identify visually grounded hard negatives, we first extract reliable sign--word correspondences using a pre-trained sign language retrieval model. For each video--text pair $(v_i, t_i)$, we compute the sign--word similarity matrix
\begin{equation}
E^{(i)} = v'_i \cdot (t'_i)^{\top} \in \mathbb{R}^{N \times M}.
\end{equation}
Subsequently, we normalize $E^{(i)}$ into a probability matrix via row-wise softmax:
\begin{equation}
P_{n,m}^{(i)} = \frac{\exp(E_{n,m}^{(i)} / \tau)}{\sum_{j=1}^{M} \exp(E_{n,j}^{(i)} / \tau)}.
\end{equation}
For each sign token $s_i^n$, we take the maximum alignment probability:
\begin{equation}
p_i^n = \max_{1 \le m \le M} P_{n,m}^{(i)}.
\end{equation}
Only matches with probabilities exceeding a threshold $\alpha$ are retained, yielding the set of reliable sign--word pairs:
\begin{equation}                                                                                                          
  \scalebox{0.88}{$\displaystyle                                                          
  \mathcal{M}_{\text{conf}} = \{ (s_i^n, t_i^k) \mid p_i^n > \alpha,\; k = \arg\max_m P_{n,m}^{(i)}\}.
  $}                                                                                                                        
\end{equation}
Note that this step is not intended to produce perfect alignments, but rather to conservatively filter out noisy alignments and stabilize the subsequent negative mining process.

\paragraph{Visually Similar Negative Word Mining}

Given $\mathcal{M}_{\text{conf}}$, we identify visually similar but semantically distinct signs to construct hard negative candidates.
For each $(s_i^n, t_i^k) \in \mathcal{M}_{\text{conf}}$, we search for other pairs $(s_j^m, t_j^l)$ whose sign features are visually similar by measuring:
\begin{equation}
\text{sim}(s_i^n, s_j^m) = \frac{s_i^n \cdot s_j^m}{\|s_i^n\| \|s_j^m\|}, \quad j \neq i.
\end{equation}
We retain candidates whose similarity exceeds a threshold $\beta$ and whose word tokens differ:
\begin{equation}
\mathcal{N}(t_i^k) = \{ t_j^l \mid \text{sim}(s_i^n, s_j^m) > \beta,\; t_j^l \neq t_i^k \}.
\end{equation}

By selecting negatives based on visual proximity between signs, this step reflects the constrained and structured articulation space of sign language, where visual similarity, rather than linguistic relatedness, governs confusability.

\paragraph{Hard Negative Caption Generation}

Using the mined candidate sets $\mathcal{N}$, we generate hard negative captions via keyword substitution.
For each original caption $t_i$, we identify words with non-empty $\mathcal{N}(t)$, randomly select such words, and replace them with samples drawn from their corresponding candidate sets, yielding a set of hard negatives $H_i$.

We incorporate these negatives into contrastive learning via the following objective:
\begin{equation}
  \scalebox{0.79}{$\displaystyle
  \mathcal{L}_{\textit{fine}} = -\frac{1}{B} \sum_{i=1}^{B}      \log \frac{\exp(S_{V2T}(v_i, t_i)/\tau)}                       {\sum_{x \in H_i \cup \{t_i\}} \exp(S_{V2T}(v_i, x)/\tau)}.     $}                                                          \end{equation}  
Here, the number of substituted words and hard negatives serve as design choices, and their impact is systematically analyzed in Section \ref{sec:analysis}.

\paragraph{Final Objective}

The final training objective is defined as:
\begin{equation}
\mathcal{L} = \mathcal{L}_{\textit{coarse}} + \lambda \cdot \mathcal{L}_{\textit{fine}},
\end{equation}
where $\lambda$ balances coarse-grained alignment and sign-aware hard negative supervision.
This formulation allows us to isolate the effect of SAN while preserving the original retrieval objective, enabling controlled analysis in subsequent experiments.







\section{Experiments}
\subsection{Experimental Setup}

\paragraph{Dataset and Evaluation Goal}
We conduct experiments on \textbf{PHOENIX-2014T}~\cite{DBLP:conf/cvpr/slt_ref1}, a standard benchmark for German Sign Language consisting of 7,096 training, 519 validation, and 642 test video--text pairs collected from TV weather forecast broadcasts. Beyond standard retrieval accuracy, our goal is to evaluate a model’s ability to resolve \emph{visually confusable signs}, which frequently arise in sign language due to its constrained articulatory space, yet are not explicitly captured by existing benchmarks.

\paragraph{Coarse-grained and Fine-grained Evaluation}
We consider two complementary evaluation settings.
\textbf{Coarse-grained retrieval} utilizes the original test split and measures overall video--text matching performance under standard conditions, but does not probe whether a model can distinguish signs that differ only in subtle motion details. 
To explicitly assess this capability, we introduce a \textbf{fine-grained evaluation setting}, designed as a diagnostic \emph{stress test} on visually ambiguous regions of the dataset.

\paragraph{Fine-grained Stress Test Construction}
The fine-grained test set is constructed via \emph{single-word substitution}, producing minimally perturbed but semantically incorrect captions. 
Target words are restricted to those exhibiting visual confusability—specifically, words whose corresponding signs have at least one visually similar counterpart (cosine similarity $> \beta$ in the sign embedding space) among semantically distinct signs. 
Since existing benchmarks lack ground-truth annotations for sign-level visual similarity, we use sign embedding similarity as a proxy. 
Once identified, the same target words are used across all methods, ensuring a fair comparison.
All target words are selected exclusively from the test split to prevent train--test leakage.

\paragraph{Controlled Comparison and Training Protocol}
All methods are evaluated under a strictly controlled environment, ensuring that the same target words and substitution positions are used across all models. For each target word, we generate hard negative captions by replacing it with negative candidate words produced by the SAN and the comparative language models. In total, each method generates 10 hard negative captions per sentence, resulting in 40 negatives for each original caption. The fine-grained evaluation assesses how accurately the model retrieves the original caption from among these hard negative captions. During training, we employ multi-word substitutions to improve optimization stability, while fine-grained evaluation uses single-word substitutions to create more challenging minimally perturbed queries. 
The impact of negative difficulty and substitution granularity is further analyzed in Section~\ref{sec:analysis}.

\paragraph{Evaluation Metrics}
We report Recall at rank $k$ (R@$k$) for $k \in \{1,5,10\}$ and Mean Reciprocal Rank (MRR).

\paragraph{Implementation Details}
We adopt GFSLT-VLP~\cite{DBLP:conf/iccv/slt_ref3} and CiCo~\cite{DBLP:conf/cvpr/cico} as our base architectures and train all models with the \textbf{Cross-Lingual Contrastive Learning (CLCL)} objective~\cite{DBLP:conf/cvpr/cico}. All models are trained for 100 epochs using SGD with an initial learning rate of $1\times10^{-2}$ and a cosine learning rate scheduler. We use a batch size of 32 for GFSLT-VLP and 256 for CiCo. For the SAN framework, we set  $\alpha=0.7$ and $\beta=0.7$ for mining negatives in both training and evaluation, with the balancing weight $\lambda=0.4$ applied during training. We utilize a trained GFSLT-VLP-based sign language retrieval model for the SAN framework.

\paragraph{Language Models for Negative Word Mining}
Previous vision-language retrieval studies have demonstrated the effectiveness of using language models to generate hard negative captions for fine-grained visual understanding~\cite{DBLP:conf/cvpr/ranking_clip, DBLP:conf/iccv/verb_in_action, DBLP:conf/cvpr/teaching_structured_clip}. Following this line of research, we employ \textbf{RoBERTa}~\cite{DBLP:journals/corr/roBERTa} and \textbf{GPT-4o-mini}~\cite{DBLP:journals/corr/abs-2303-08774} as comparative language models for hard negative generation. We also incorporate \textbf{FastText} \cite{DBLP:conf/lrec/MikolovGBPJ18}, motivated by its use in NLA-SLR \cite{DBLP:conf/cvpr/slr_ref_2} to enhance discrimination among visually similar signs in sign language recognition.

\begin{table*}[t]
\centering
\resizebox{\textwidth}{!}{
\begin{tabular}{l l | cccc | cccc | cccc}
\hline
\textbf{Model} 
& \textbf{Hard Negatives}
& \multicolumn{4}{c|}{\textbf{Fine-grained ($\uparrow$)}} 
& \multicolumn{8}{c}{\textbf{Coarse-grained ($\uparrow$)}} \\
\cline{3-14}
& 
& \multicolumn{4}{c|}{\textbf{V2T}}
& \multicolumn{4}{c|}{\textbf{T2V}}  
& \multicolumn{4}{c}{\textbf{V2T}} \\
\cline{3-14}
&
& R@1 & R@5 & R@10 & MRR
& R@1 & R@5 & R@10 & MRR
& R@1 & R@5 & R@10 & MRR\\
\hline

Cico
& $\varnothing$ 
& 17.9 & 55.3 & 79.1 & 35.0
& \textbf{69.2} & {87.2} & \textbf{92.2} & \textbf{77.3}
& \textbf{70.1} & \textbf{87.7} & \textbf{92.9} & \textbf{78.2} \\
 & FastText
& {33.8} & \textbf{76.8} & 92.1 & {51.8} 
& 67.3 & 86.1 & 91.6 & 75.4 
& 67.0 & 86.0 & 91.1 & 75.4 \\
 & RoBERTa
& 25.6 & 68.2 & 88.9 & 43.2 
& 67.1 & 86.5 & {91.9} & 75.7 
& 63.6 & 84.7 & 89.4 & 73.1 \\
 & GPT4o-mini
& 30.7 & {76.3} & \textbf{92.7} & 49.1 
& 67.5 & 87.1 & 91.6 & 75.9 
& 67.6 & 86.1 & 91.3 & 75.8 \\
 & SAN (Ours)
& \textbf{39.4} & 75.4 & {92.5} & \textbf{54.4}
& {68.1} & \textbf{87.4} & 91.7 & {76.6} 
& {67.8} & {87.4} & {91.7} & {76.2} \\
\hline

GFSLT-VLP 
& $\varnothing$ 
& 16.8 & 53.1 & 78.0 & 33.9 
& 67.9 & 88.4 & 93.8 & 77.5 
& \textbf{69.4} & {88.7} & \textbf{93.3} & \textbf{77.9}  \\
 & FastText
& {43.8} & {85.6} & \textbf{96.7} & {61.3} 
& 68.6 & \textbf{90.1} & 93.2 & 77.7 
& 64.3 & 84.8 & 89.9 & 73.4 \\
 & RoBERTa
& 29.5 & 73.9 & 95.0 & 47.9 
& 68.0 & {89.6} & 93.9 & 77.1 
& 65.2 & 86.7 & 90.8 & 74.3 \\
 & GPT4o-mini
& 37.7 & 85.3 & {96.4} & 56.4 
& \textbf{70.9} & 89.4 & {94.1} & \textbf{79.1} 
& 66.6 & \textbf{89.1} & {92.1} & {76.2} \\ 
 & SAN (Ours)
& \textbf{49.1} & \textbf{85.9} & 94.9 & \textbf{64.1} 
& {70.2} & 89.3 & \textbf{94.4} & {78.7} 
& {67.4} & 85.4 & 90.5 & 75.5 \\
\hline
\end{tabular}
}
\caption{Retrieval performance on coarse-grained and fine-grained evaluation settings on PHOENIX-2014T. \textbf{Bold} indicates the best performance.}
\label{main_table}
\vspace{-0.2in}
\end{table*}

\subsection{Retrieval Results}
In Table~\ref{main_table}, we compare coarse- and fine-grained retrieval performance under different hard-negative mining strategies.

\paragraph{Unlocking Fine-grained Capabilities with SAN}
We first investigate the impact of our proposed SAN strategy to validate its effectiveness in enhancing fine-grained discrimination.
As shown in Table~\ref{main_table}, standard retrieval models struggle significantly with fine-grained queries, achieving only 17.9\% and 16.8\% in V2T R@1 for CiCo and GFSLT-VLP, respectively. However, incorporating our proposed SAN strategy leads to substantial performance gains across both architectures.
Specifically, applying SAN to CiCo boosts the fine-grained V2T R@1 from 17.9\% to \textbf{39.4\%}, a remarkable absolute improvement of \textbf{+21.5\%}. The impact is even more pronounced with GFSLT-VLP, where SAN nearly triples the baseline performance, jumping from 16.8\% to \textbf{49.1\%} (\textbf{+32.3\%}).

These results suggest that the suboptimal fine-grained performance of existing baselines stems not from insufficient model capacity, but from the lack of exposure to ambiguous instances during training. This highlights the necessity of explicitly learning from confusing negatives: SAN forces the model to resolve these ambiguities, thereby obtaining the discriminative features required to distinguish subtle sign variations.

\paragraph{Comparison with Text-based Mining Strategies} 
We further compare SAN with text-based hard negative mining strategies, including FastText, RoBERTa, and GPT-4o-mini. As illustrated in Table~\ref{main_table}, SAN consistently outperforms all text-driven baselines on the fine-grained test set.
On the GFSLT-VLP backbone, our method achieves a V2T R@1 of \textbf{49.1\%}, surpassing the strongest text-based competitor, FastText (43.8\%), by a clear margin. Notably, leveraging a state-of-the-art LLM (GPT-4o-mini) yields only 37.7\%, falling short of our visual-aware approach by \textbf{11.4} percentage points.
A similar trend is observed with CiCo, where SAN (39.4\%) significantly outperforms GPT-4o-mini (30.7\%) and RoBERTa (25.6\%).
These results underscore a critical insight: \emph{linguistic hardness does not equate to visual hardness} in sign language. While text-based methods generate semantically related negatives (e.g., synonyms), these words often look visually distinct when signed. In contrast, SAN directly mines negatives based on visual similarity, providing much more effective supervision for distinguishing subtle sign variations. 

\paragraph{Robustness on Coarse-grained Retrieval} Enhancing fine-grained discrimination often entails a trade-off, potentially degrading performance on standard coarse-grained benchmarks due to the distortion of global semantic alignment~\cite{DBLP:conf/accv/beyond}. However, our results indicate that SAN offers a significantly more favorable trade-off compared to text-based mining strategies. On the GFSLT-VLP backbone, text-based methods notably degrade coarse-grained V2T performance (e.g., FastText drops R@1 from 69.4\% to 64.3\%). In contrast, SAN maintains a robust performance of 67.4\%, mitigating the degradation. A similar trend is observed with CiCo, where SAN exhibits the smallest performance drop among all mining-based methods. This suggests that visual-aware negatives not only refine detailed visual understanding but can also effectively maintain coarse-grained alignment, whereas linguistic negatives often introduce noise that disrupts standard retrieval.


\section{Analysis}
\begin{figure}[!t]
      \centering
      \begin{subfigure}{0.22\textwidth}
          \centering
          \includegraphics[width=\textwidth]{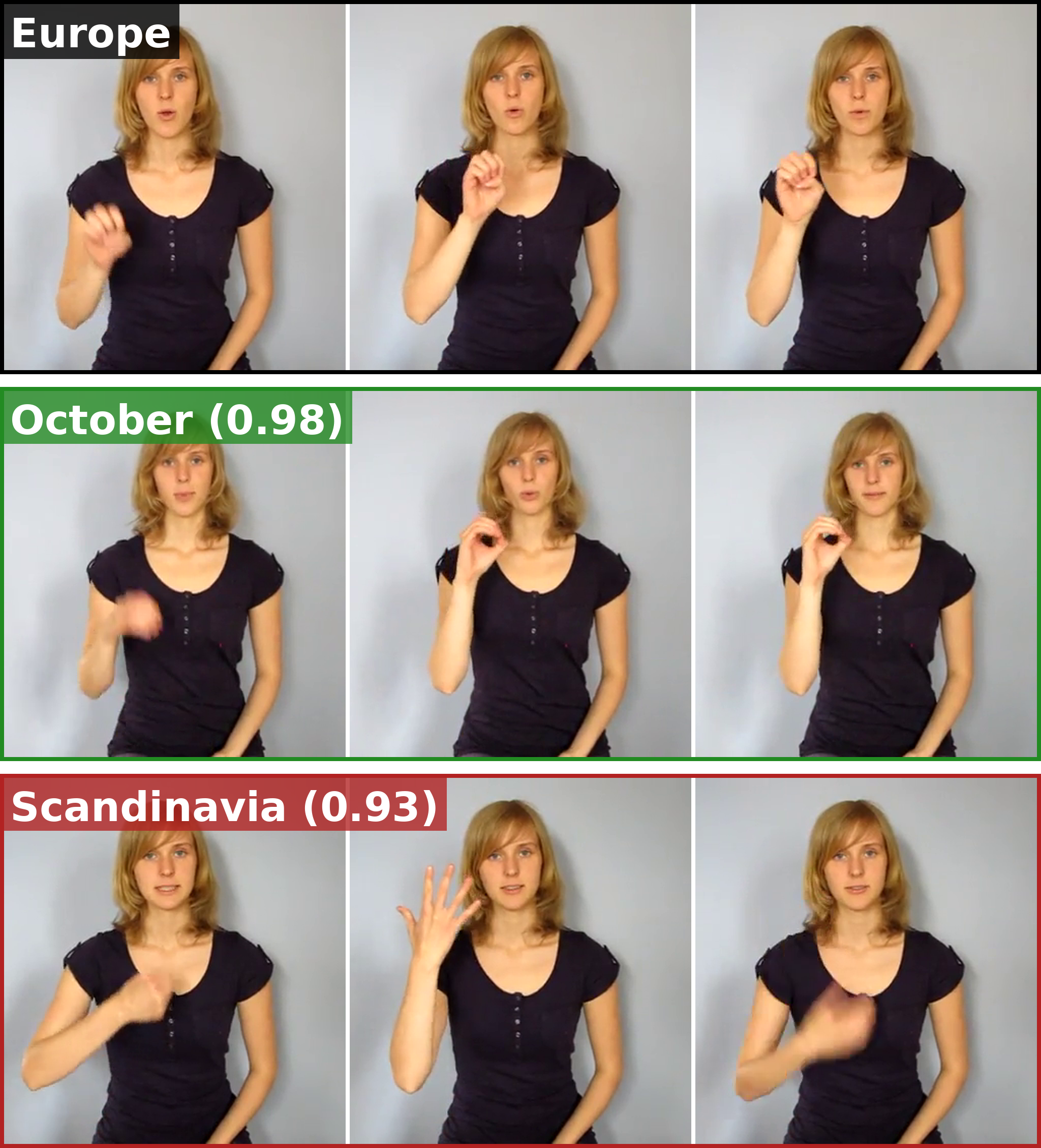}
          \caption{Europa}
          \label{fig:Europa}
      \end{subfigure}
      \hfill
      \begin{subfigure}{0.22\textwidth}
          \centering
          \includegraphics[width=\textwidth]{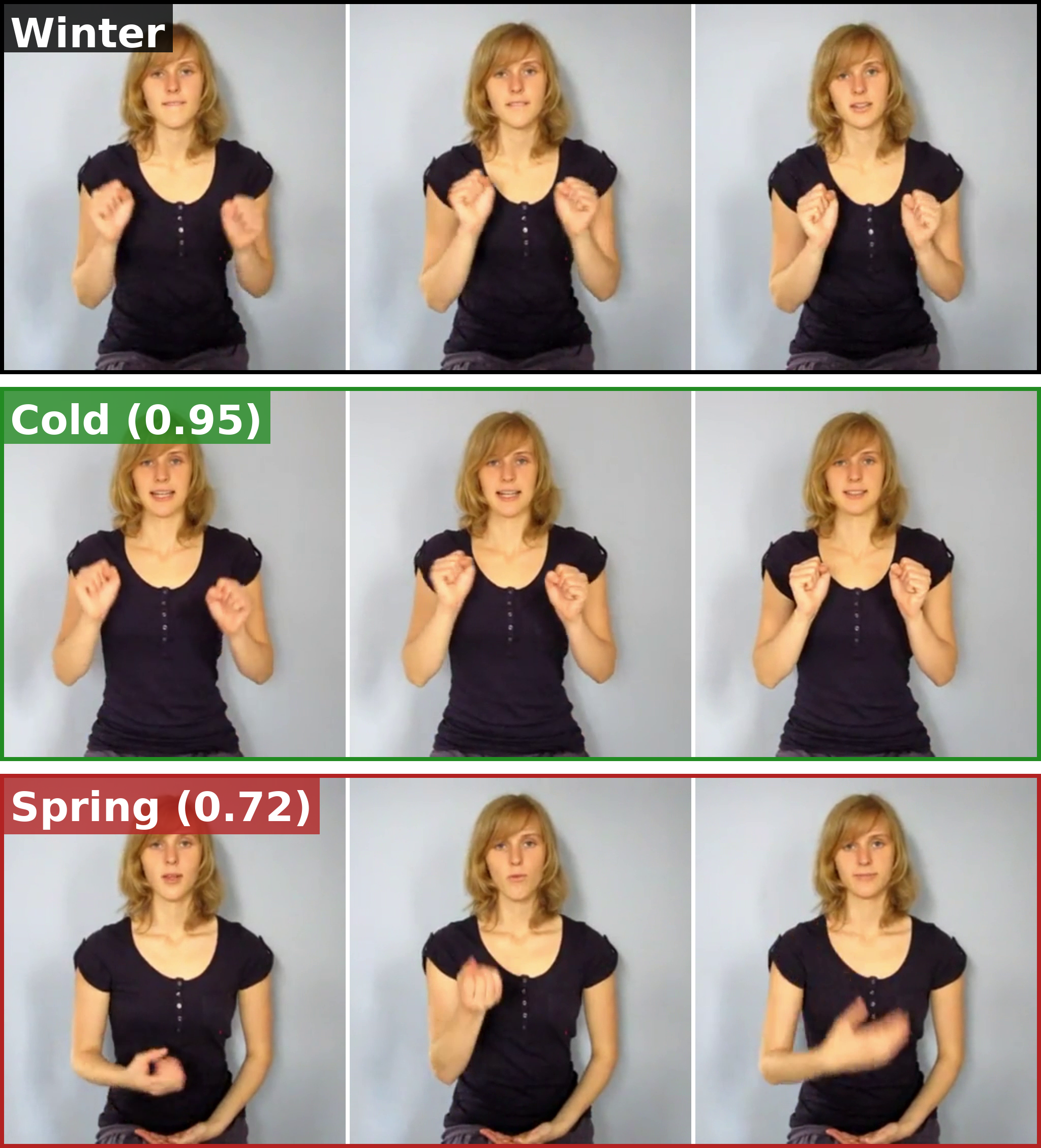}
          \caption{Winter}
          \label{fig:winter}
      \end{subfigure}
      \begin{subfigure}{0.22\textwidth}
          \centering
          \includegraphics[width=\textwidth]{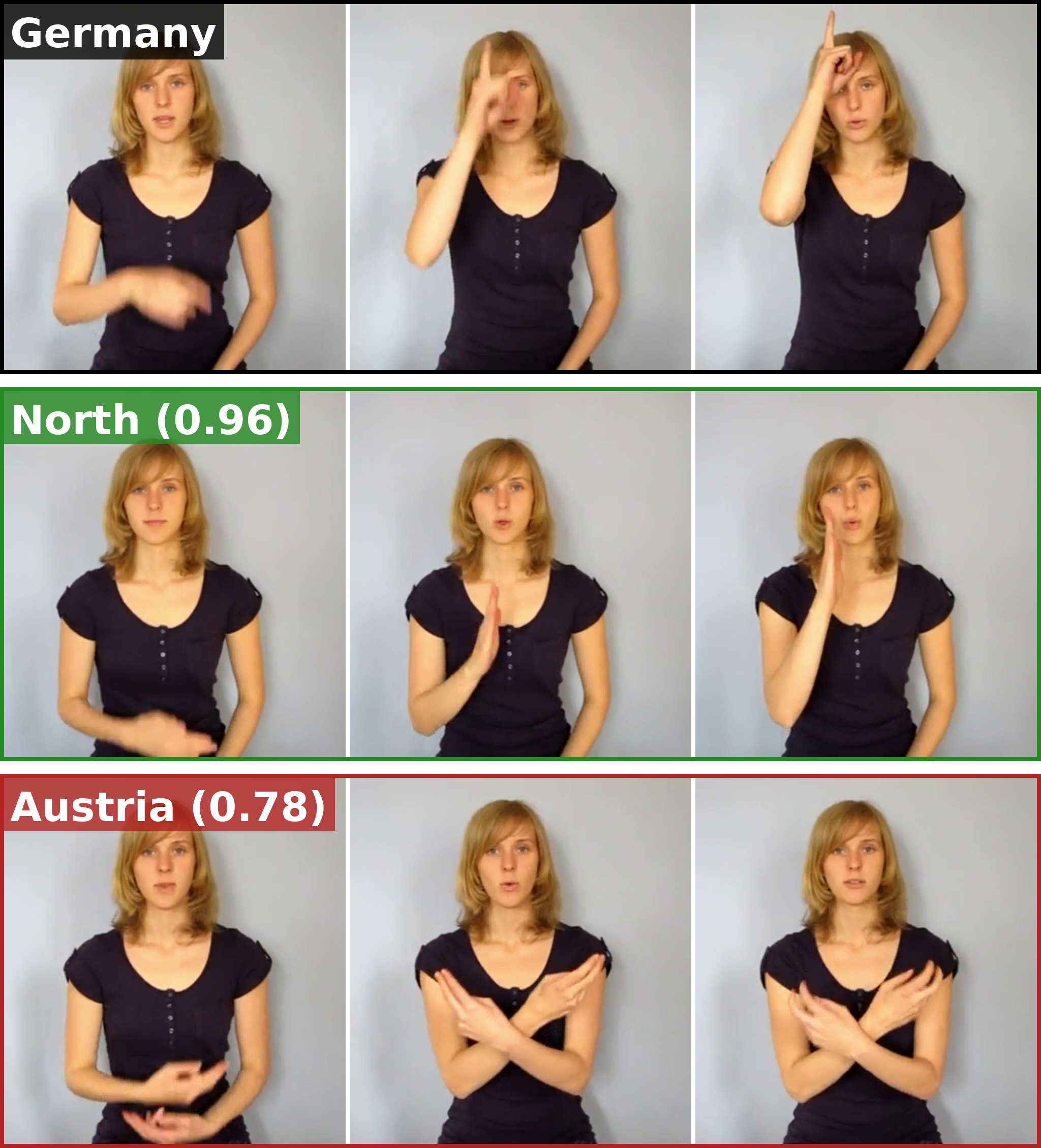}
          \caption{Deutschland}
          \label{fig:deutschland}
      \end{subfigure}
      \hfill
      \begin{subfigure}{0.22\textwidth}
          \centering
          \includegraphics[width=\textwidth]{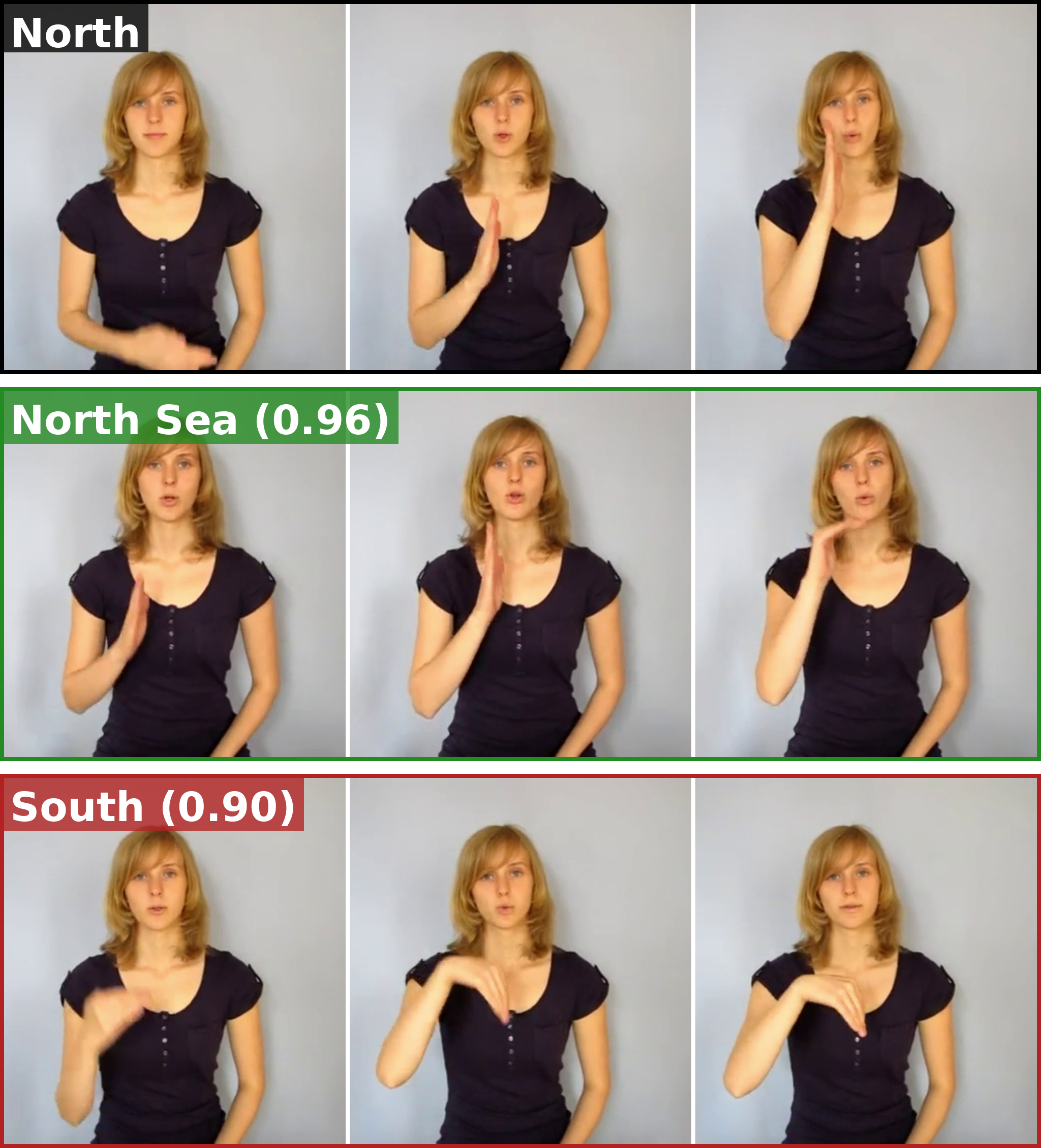}
          \caption{Norden}
          \label{fig:norden}
      \end{subfigure}

      \caption{Qualitative comparison of mined hard negative words.
For each example, the first row shows the sign of the target word, 
the second row shows the negative word mined by \textcolor{green}{SAN}, 
and the third row shows a negative mined by  \textcolor{red}{text-based} methods. \textcolor{green}{SAN} retrieves visually similar signs with subtle motion differences, whereas  \textcolor{red}{text-based} methods select linguistically plausible but visually distinct negatives. The scores indicate the similarity of the sign video features extracted by the pretrained I3D model.}
      \label{fig:word_retrieval_comparison}
      \vspace{-0.1in}
  \end{figure}

\paragraph{Visual Comparison of Mined Negative Words}
\label{sec:analysis}
Figure~\ref{fig:word_retrieval_comparison} provides a qualitative comparison of the mined negative words in terms of signing motion. In each example, the first row shows the sign corresponding to the original word, the second row shows the negative word mined by SAN (ours), which other methods failed to mine, and the third row shows a negative word mined by text-based methods. Across all cases, SAN consistently mines negatives whose signs exhibit highly similar handshape, location, and movement to the original sign (e.g., Europe $\rightarrow$ October, Winter $\rightarrow$ Cold, Germany $\rightarrow$ North, North $\rightarrow$ North Sea), resulting in subtle motion differences that are difficult to distinguish visually. In contrast, negatives mined by text-based methods (e.g., Scandinavia, Spring, Austria, South) often involve clearly different articulations, making them less visually confusable despite being linguistically plausible substitutions. Overall, these examples suggest that SAN can mine true hard negatives—negatives that are highly confusable in sign motion.

\begin{figure}[t]
    \centering
    \hspace{-0.1\linewidth}
    \includegraphics[width=\columnwidth]{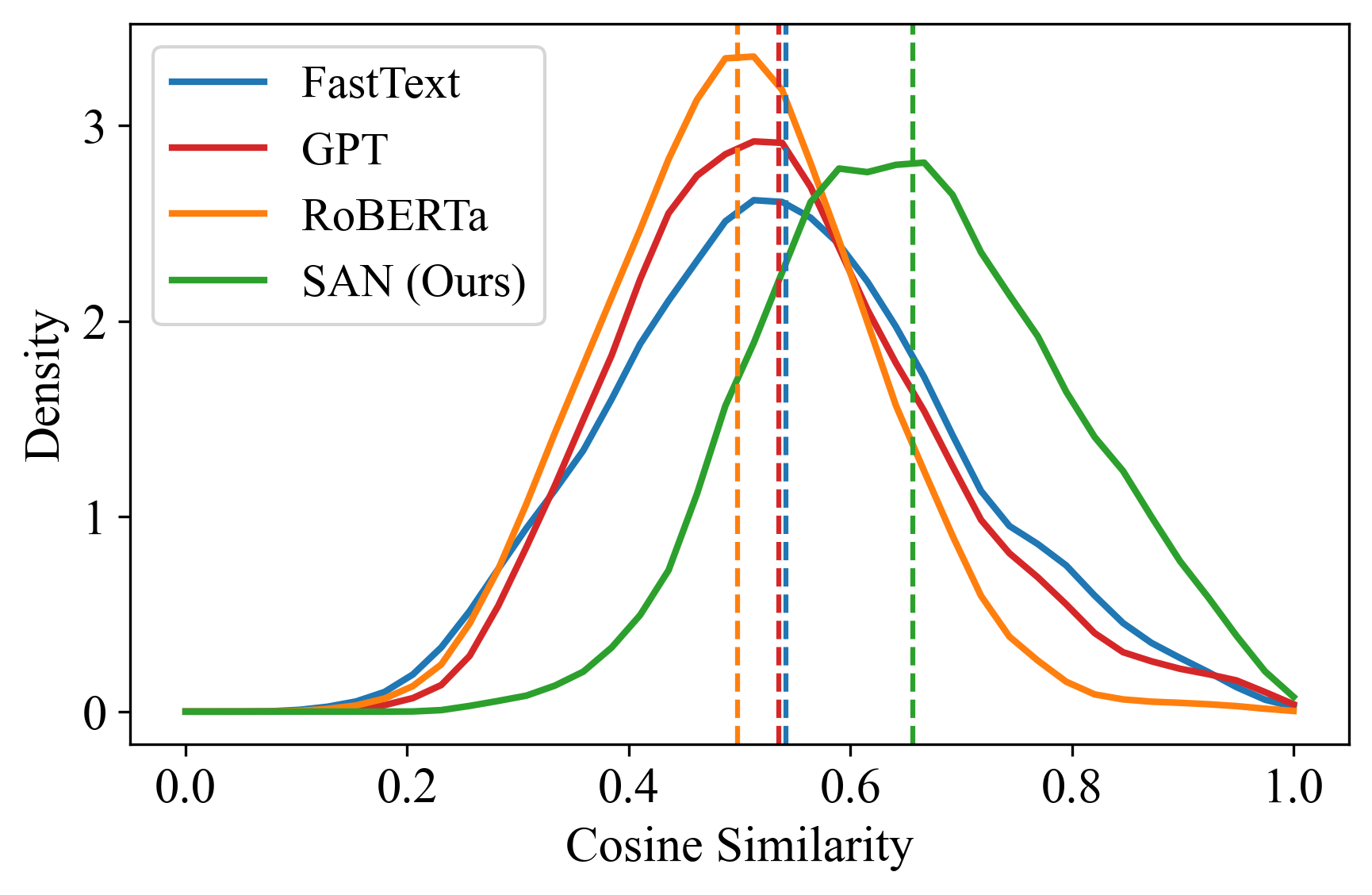}
    \caption{Distribution of cosine similarities between sign prototypes of target and negative words. The distribution of SAN is shifted toward higher similarity values, demonstrating better coverage of visually confusable regions than text-based methods.}
    \label{fig:cosine_sim_distribution}
    \vspace{-0.1in}
\end{figure}

\begin{figure}[t]
    \centering
    \includegraphics[width=\columnwidth]{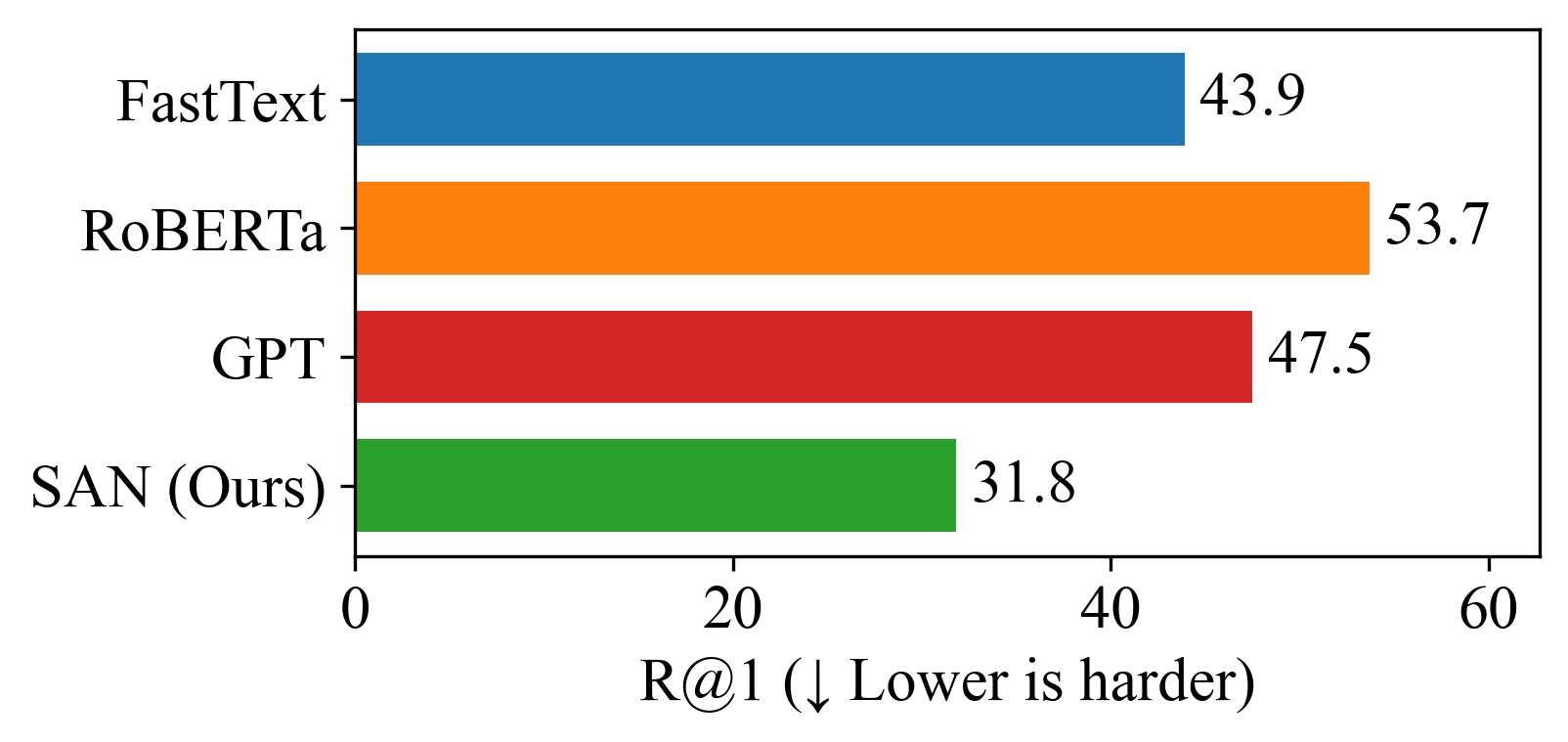}
    \caption{Comparison of the hardness of generated negatives. Lower retrieval performance implies higher difficulty.}
    \label{fig:hard_test_analysis}
    \vspace{-0.1in}
\end{figure}

\paragraph{Quantitative Analysis of Negative Distribution Mismatch}
We further analyze the degree of visual confusability of mined negative words. To this end, we establish word-to-sign alignments via the trained CiCo~\cite{DBLP:conf/cvpr/cico} model and generate sign prototypes by averaging the corresponding sign features for each word. 
Figure~\ref{fig:cosine_sim_distribution} presents the cosine similarity distributions between negative words and their corresponding target words for each mining strategy. 
Notably, negatives mined by SAN exhibit a clear shift toward higher cosine similarity values compared to those mined by FastText, RoBERTa, and GPT-4o-mini. This shift suggests that SAN is more likely to sample negatives from the visually confusable region, thereby mitigating the negative distribution mismatch problem. By contrast, text-based substitutions fail to adequately cover this region, as they prioritize linguistic plausibility over visual similarity.

\begin{table}[t] 
    \centering
    \small 
    \setlength{\tabcolsep}{3pt} 
    \renewcommand{\arraystretch}{1.1}
    
    \begin{tabular}{c c cc @{\hspace{15pt}} c c cc} 
        \toprule
        \multirow{2}{*}{$N_{\text{swap}}$} & \textbf{FG} ($\uparrow$) & \multicolumn{2}{c}{\textbf{CG} ($\uparrow$)} & 
        \multirow{2}{*}{$N_{\text{hard}}$} & \textbf{FG} ($\uparrow$) & \multicolumn{2}{c}{\textbf{CG} ($\uparrow$)} \\
        \cmidrule(lr){2-2} \cmidrule(lr){3-4} \cmidrule(lr){6-6} \cmidrule(lr){7-8}
        & V2T & T2V & V2T & & V2T & T2V & V2T \\
        \midrule
        $\varnothing$ & 17.9 & 69.2 & 70.1 & $\varnothing$ & 17.9 & 69.2 & 70.1 \\
        1             & 38.6 & 65.0 & 63.1 & 3             & 35.8 & 66.7 & 68.2 \\
        2             & 39.4 & 68.1 & 67.8 & 5             & 39.4 & 68.1 & 67.8 \\
        3             & 34.9 & 68.1 & 69.9 & 7             & 39.7 & 67.6 & 67.3 \\
        \bottomrule
    \end{tabular}
    \caption{Ablation study on the number of swapped words $N_{\text{swap}}$ (left) and the number of hard negatives $N_{\text{hard}}$ (right). FG and CG denote fine-grained and coarse-grained, respectively.}
    \label{tab:ablations}
    \vspace{-0.1in}
\end{table}






\paragraph{Difficulty of Generated Hard Negative Captions}

To validate the intrinsic quality of the mined negatives, we evaluate the retrieval performance on the specific hard-negative sets generated by each method, rather than a merged set. The goal of this experiment is to quantify the discriminative challenge posed by the generated negatives. As illustrated in Figure~\ref{fig:hard_test_analysis}, we observe a substantial disparity in retrieval performance depending on the mining strategy. While the model maintains high accuracy on negatives generated by language models (e.g., RoBERTa: 53.7\%, GPT-4o-mini: 47.5\%), its performance drops sharply to 31.8\% on the SAN-generated set. Crucially, this performance drop does not indicate model failure, but rather confirms the high quality of our hard negatives. Unlike text-based baselines that produce easily distinguishable negatives, SAN successfully mines visually ambiguous instances that effectively challenge the model’s discriminative capabilities.



\begin{figure}
    \centering
    \begin{minipage}[t]{0.48\columnwidth}
        \centering
        \includegraphics[width=\linewidth]{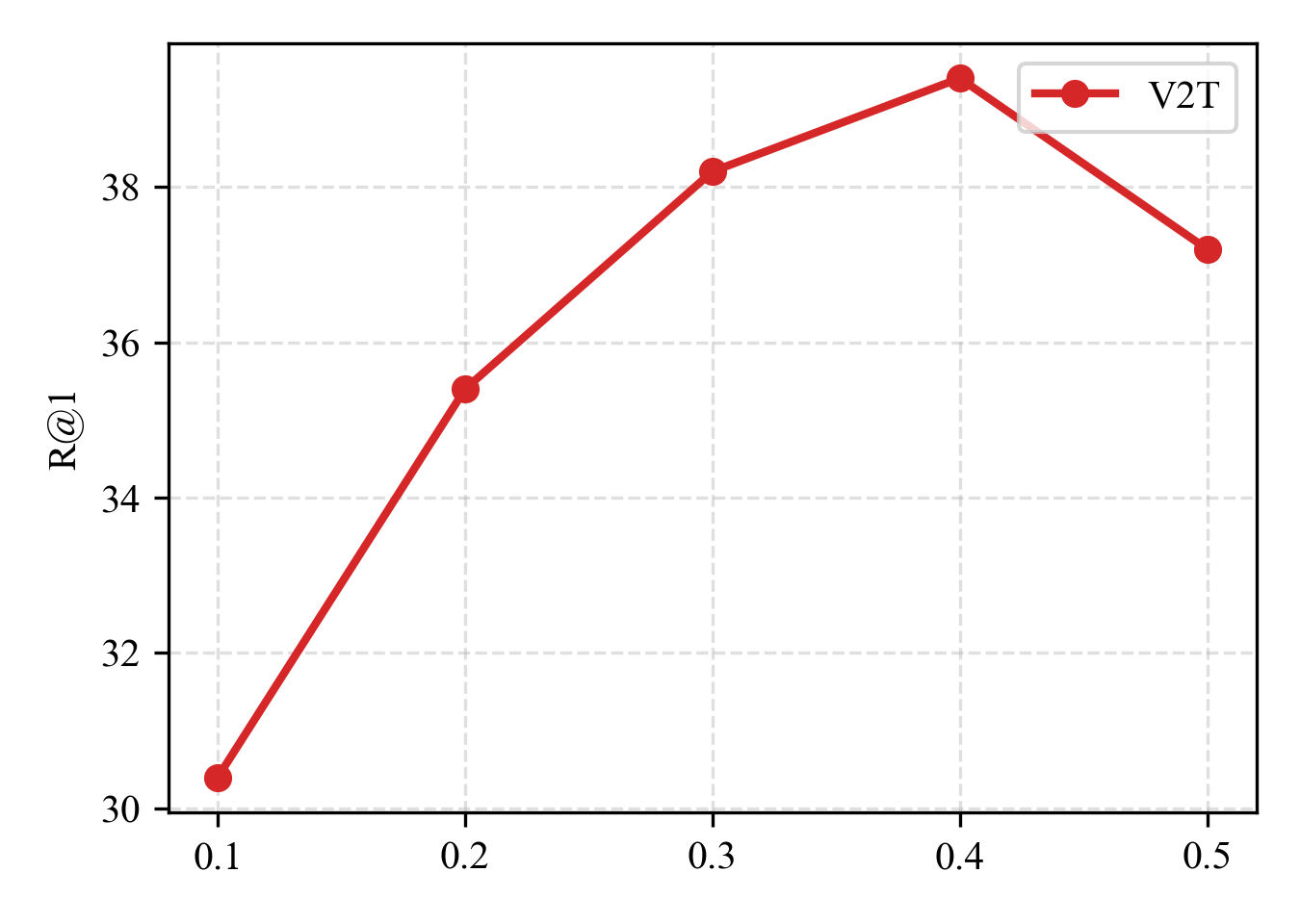}
        \caption*{(a) Fine-grained}
    \end{minipage}
    \hfill
    \begin{minipage}[t]{0.48\columnwidth}
        \centering
        \includegraphics[width=\linewidth]{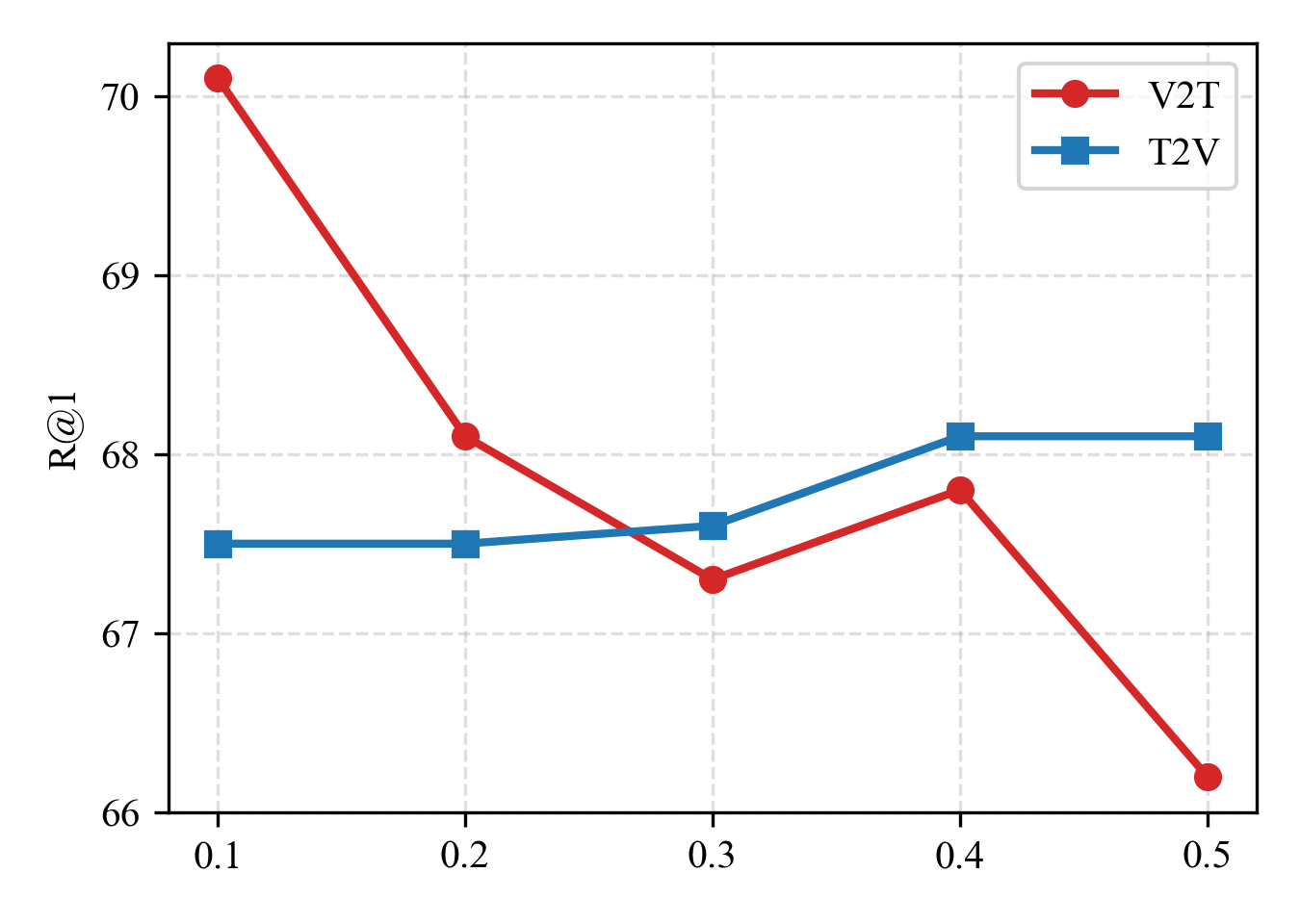}
        \caption*{(b) Coarse-grained}
    \end{minipage}
    \caption{Effect of $\lambda$ on fine-grained and coarse-grained retrieval performance.}
    \label{fig:lambda_ablation}
    \vspace{-0.1in}
\end{figure}

\paragraph{Impact of Number of Swapped Words}
In Table~\ref{tab:ablations} (Left), we show the impact of $N_{\text{swap}}$, the number of words replaced when generating hard-negative captions. Replacing only a single word yields a substantial improvement in fine-grained retrieval, but also leads to a noticeable drop in coarse-grained performance. 
This suggests that extremely hard negatives formed by minimal perturbations can over-constrain the model and disrupt the global representation space. As more words are swapped, coarse-grained performance is better preserved, while fine-grained performance gradually decreases, since the negatives become less targeted to specific visually confusable sign units. Overall, these results highlight the importance of balancing fine-grained discrimination and global robustness, and we select $N_{\text{swap}}=2$ as it achieves the best trade-off between the two.

\paragraph{Impact of Number of Hard Negatives}
We further examine the effect of $N_{\text{hard}}$, the number of hard-negative captions used per training instance. As shown in Table~\ref{tab:ablations} (Right),
increasing $N_{\text{hard}}$ consistently improves fine-grained performance, as the model is exposed to a richer set of visually confusing alternatives. However, using too many hard negatives gradually degrades coarse-grained retrieval, indicating a bias toward local discrimination at the expense of global alignment. Conversely, using too few hard negatives provides insufficient fine-grained supervision, leading to suboptimal fine-grained performance. We therefore adopt $N_{\text{hard}}=5$, which yields near-maximal fine-grained gains while limiting the drop in coarse-grained performance.

\paragraph{Sensitivity to Loss Weighting}
We analyze the effect of the loss weight $\lambda$, which controls the contribution of the hard-negative contrastive objective. As shown in Figure~\ref{fig:lambda_ablation}, fine-grained retrieval performance consistently improves as $\lambda$ increases, confirming that prioritizing these negatives effectively sharpens the model's discriminative power. However, overly large $\lambda$ leads to a noticeable degradation in coarse-grained retrieval, suggesting that excessive emphasis on hard negatives can distort the global video–text embedding structure required for general retrieval. Based on this trade-off, we select $\lambda=0.4$, which provides strong fine-grained gains while keeping coarse-grained performance largely stable.

\section{Conclusion}

In this paper, we identify a critical limitation of current sign language retrieval models: their inability to distinguish visually similar signs in fine-grained retrieval scenarios. We show that this limitation does not stem from insufficient model capacity, but from a negative distribution mismatch in contrastive learning, where training supervision fails to reflect the visual ambiguity inherent in sign language. To address this issue, we propose Sign-Aware Hard Negative Mining (SAN), which redefines hard negatives based on visual confusability in the sign embedding space rather than linguistic similarity. Experiments on PHOENIX-2014T demonstrate that SAN substantially improves fine-grained retrieval performance while preserving coarse-grained accuracy, outperforming language-model–based negative mining strategies. These results highlight the importance of shifting negative supervision signals from linguistic similarity to visual similarity of signs, suggesting that visually grounded negative mining is essential for fine-grained sign language retrieval.

\section*{Limitations}
Our experiments are limited to the PHOENIX-2014T dataset within a single domain, and the generalizability of SAN across other sign languages and datasets remains to be validated. However, we believe that PHOENIX-2014T provides a particularly suitable testbed for studying fine-grained sign language retrieval, as its weather-forecast domain naturally exhibits a dense distribution of visually similar signs (e.g., regional names, numbers, and weather-related terms), creating a concentrated environment for evaluating visual confusability. Furthermore, although we employed POS-filtering to ensure grammatical consistency during fine-grained evaluation, some linguistic noise may still persist in the generated captions. Additionally, the current framework relies on a fixed visual similarity threshold ($\beta$), which may not account for the varying degrees of ambiguity across different sign classes; thus, a dynamic or adaptive thresholding mechanism is needed. Lastly, while SAN enhances fine-grained discrimination, it can lead to a slight performance trade-off in coarse-grained retrieval, suggesting the need for balanced optimization strategies to maintain overall retrieval robustness.

\section{Discussion}
\noindent
Phonological features of signs~\cite{DBLP:journals/llc/Sandler12}, such as handshape, location, and movement, provide a principled basis for defining sign similarity and offer an alternative avenue for constructing hard negatives. In particular, dictionary-level phonological features can be used to identify \emph{minimal pairs}---sign pairs differing in only a single phonological component---which serve as a proxy for visual hardness without requiring visual data. While this is a promising direction, phonological features are discrete and static, and may not fully capture the continuous visual dynamics of actual signing that embedding-based approaches like SAN can directly model. Moreover, dictionary-based approaches are limited to standardized vocabulary, whereas vision-based mining offers broader coverage of real-world signs. Notably, SAN-mined negatives exhibit phonological similarity (Figure~\ref{fig:word_retrieval_comparison}), suggesting that the learned embedding space implicitly captures phonological structure. Integrating phonological minimal pairs as complementary signals is a promising direction for future work.

\section*{Ethics Statement}
For qualitative visualization purposes, we use sign language video examples obtained from the SignDict database\footnote{\url{https://signdict.org/}}.
According to the SignDict website, all videos are released under a Creative Commons license. We use these videos solely for academic illustration and analysis, in accordance with the stated licenses.

\section*{Acknowledgments}
This research was supported by the InnoCORE program of the Ministry of Science and
ICT(N10260002) and the Top-Tier AI Global HRD invitation program (RS-2025-25461932) supervised by the IITP(Institute for Information \& Communications Technology Planning \& Evaluation).


\bibliography{custom}

\clearpage

\appendix

\section{Appendix}
\label{sec:appendix}




\label{appendix:ablation}

\subsection{Impact of Threshold $\alpha$ and $\beta$ in SAN framework}
\begin{figure}
    \centering

    \begin{subfigure}[b]{0.48\columnwidth}
        \centering
        \includegraphics[width=\linewidth]{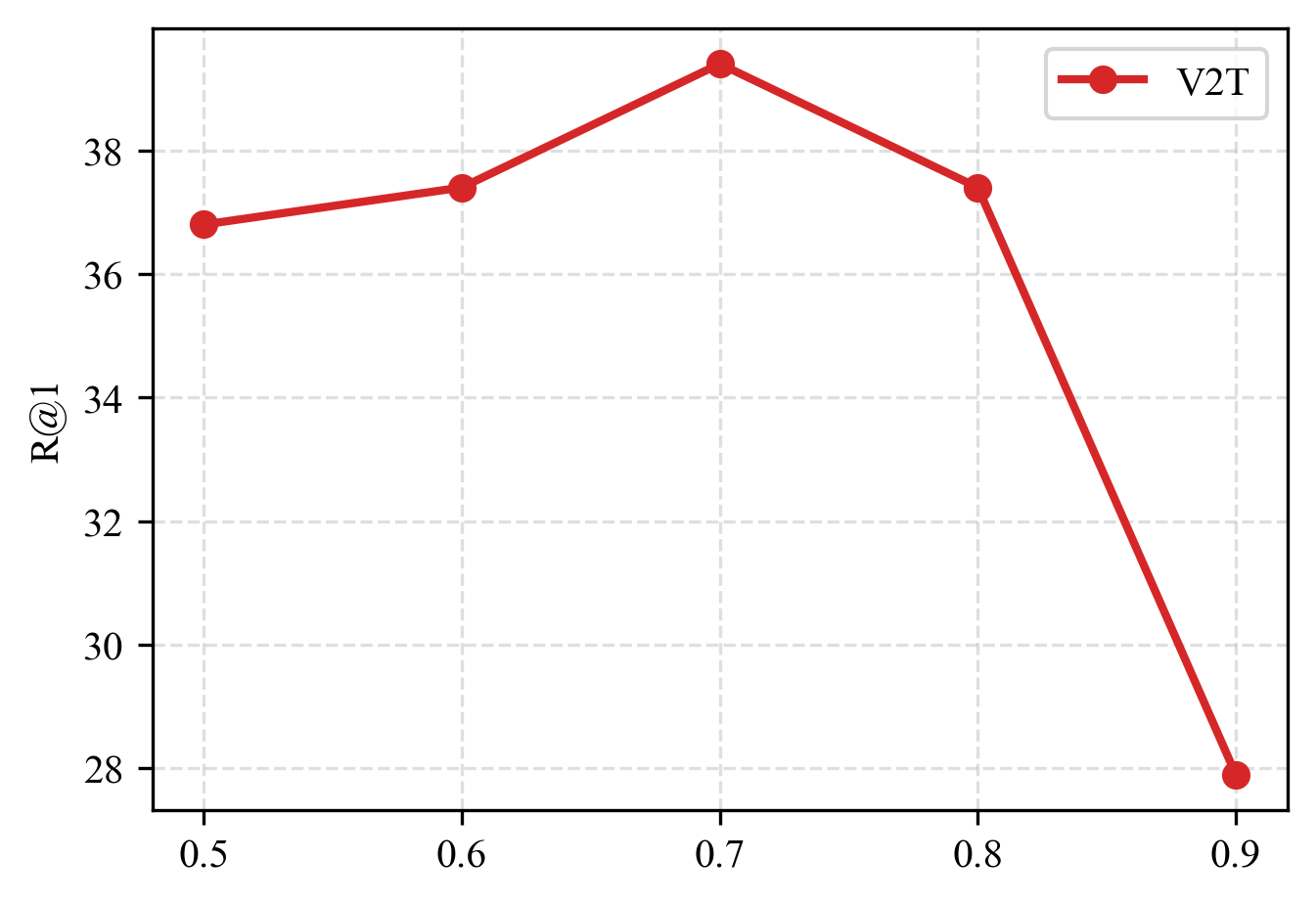}
        \label{fig:fine_alpha}
    \end{subfigure}
    \hfill
    \begin{subfigure}[b]{0.48\columnwidth}
        \centering
        \includegraphics[width=\linewidth]{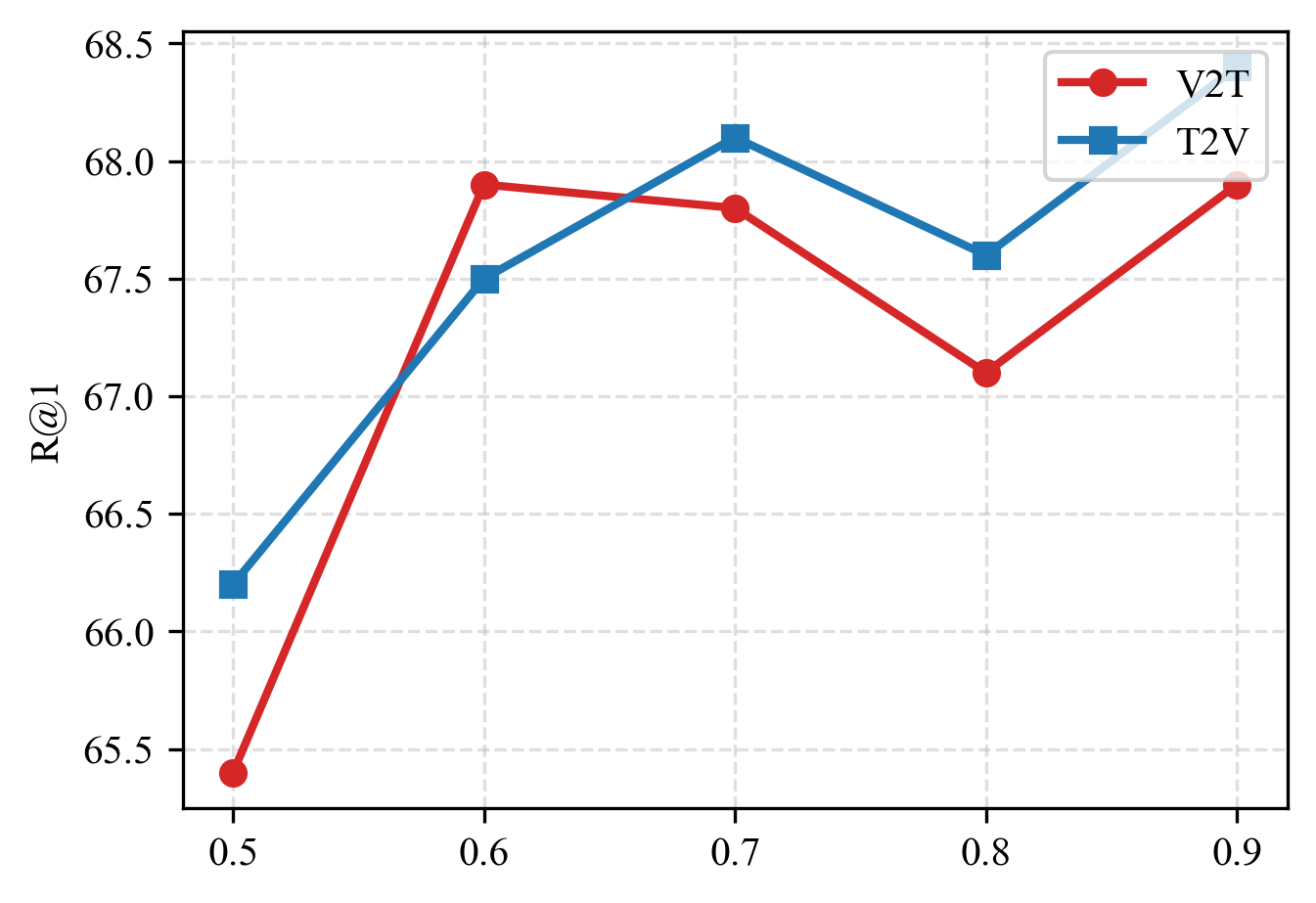}
        \label{fig:coarse_alpha}
    \end{subfigure}

    \vspace{-0.5em} 

    \begin{subfigure}[b]{0.48\columnwidth}
        \centering
        \includegraphics[width=\linewidth]{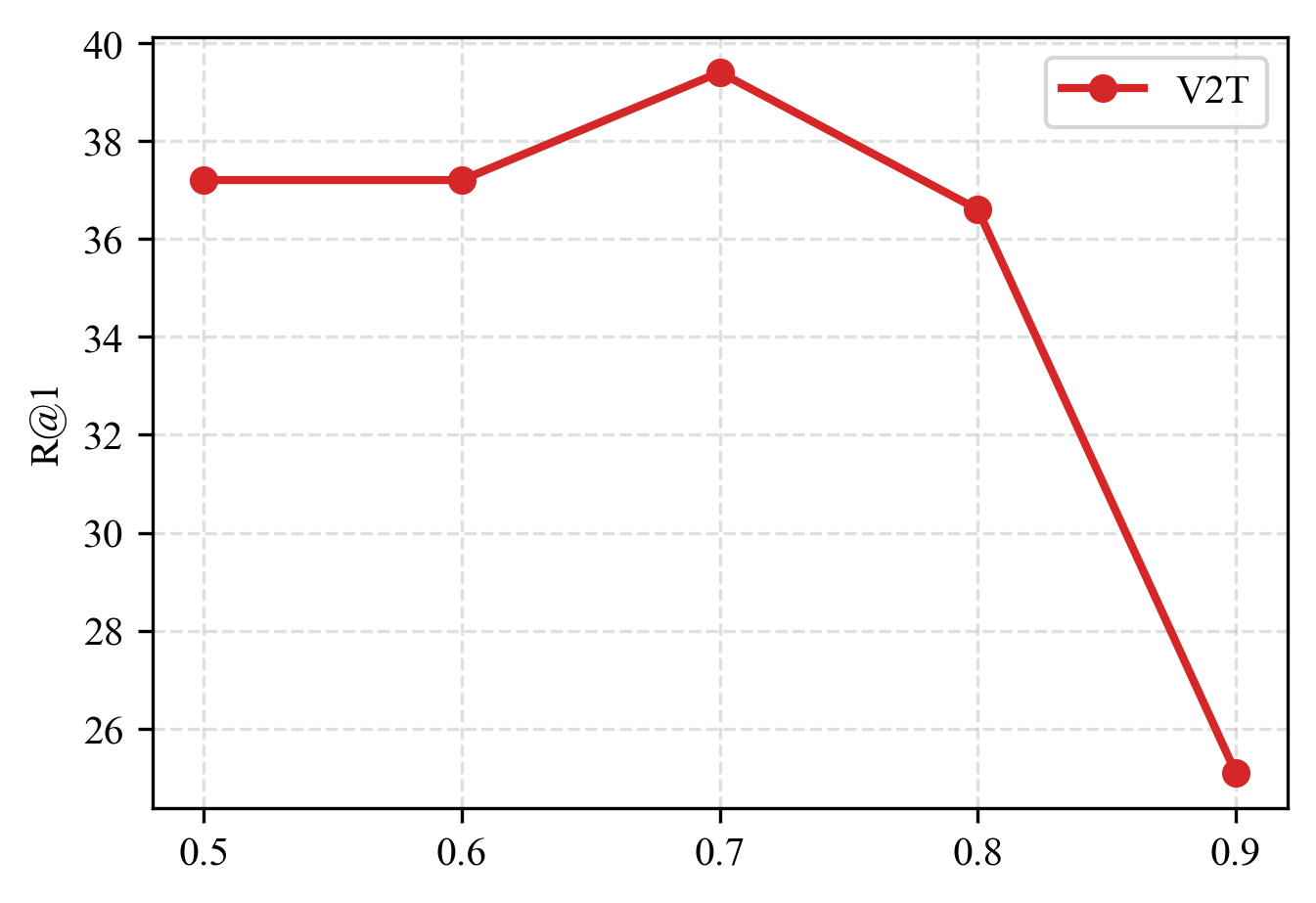}
        \caption{Fine-grained}
        \label{fig:fine_beta}
    \end{subfigure}
    \hfill
    \begin{subfigure}[b]{0.48\columnwidth}
        \centering
        \includegraphics[width=\linewidth]{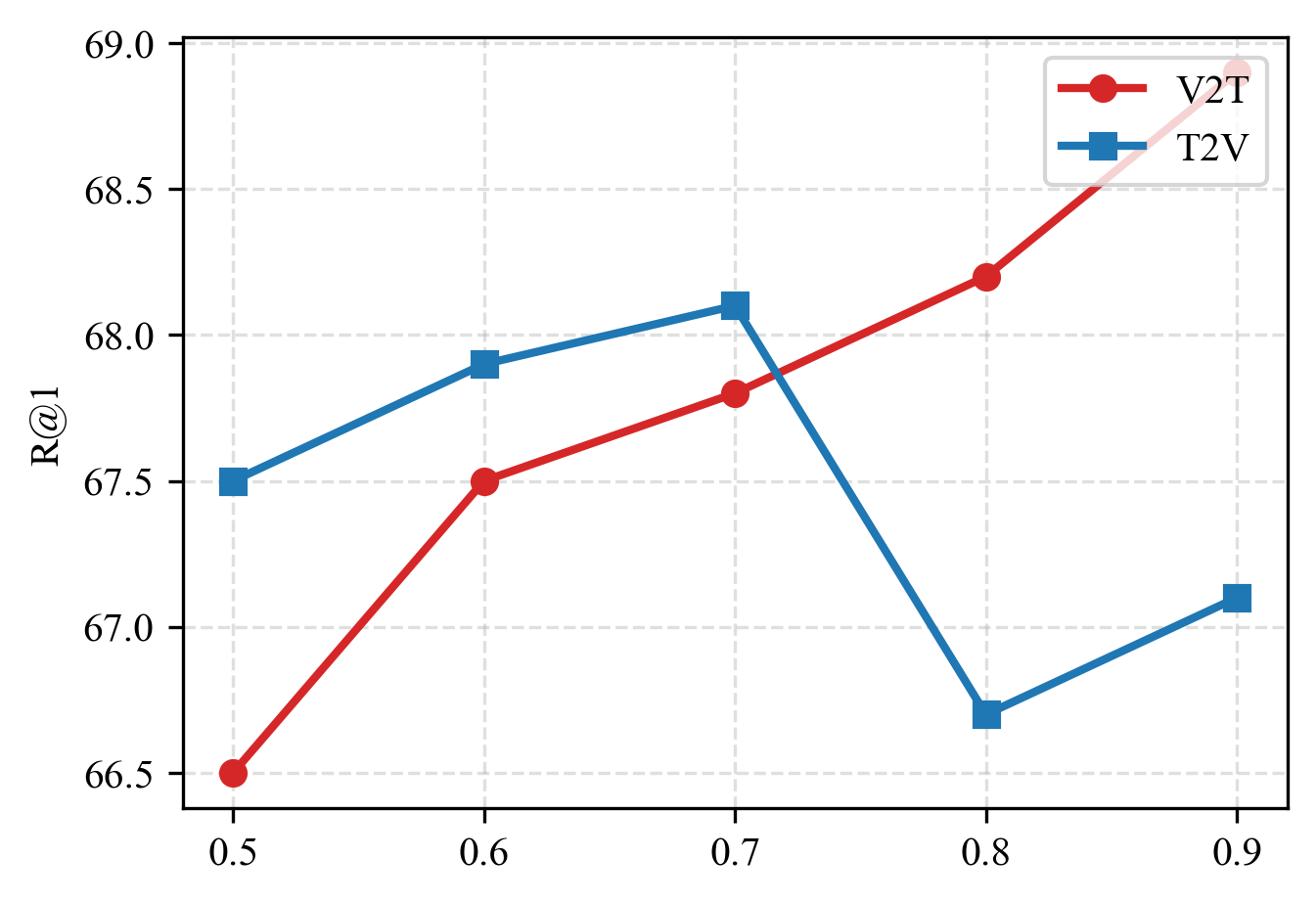}
        \caption{Coarse-grained}
        \label{fig:coarse_beta}
    \end{subfigure}

    \caption{Ablation study on hyperparameters $\alpha$ (top row) and $\beta$ (bottom row). The left column shows fine-grained retrieval results (V2T), while the right column shows coarse-grained results (V2T and T2V).}
    \label{fig:hyperparameter_ablation}
\end{figure}
We analyze the effects of two key thresholds in the SAN framework: the reliability threshold $\alpha$ used for sign--word pair mining, and the visual similarity threshold $\beta$ used for selecting visually confusable negative words. When evaluating the impact of $\alpha$, we fix $\beta$ to 0.7, and vice versa, to isolate the effect of each threshold.
Figure~\ref{fig:hyperparameter_ablation} shows that increasing $\alpha$ progressively filters out noisy sign--word alignments, leading to more stable and reliable supervision. As a result, coarse-grained retrieval performance generally improves with larger $\alpha$. However, excessively large $\alpha$ reduces fine-grained performance, as fewer sign--word pairs remain eligible for negative mining, thereby limiting the diversity and availability of informative hard negatives.
Similarly, the threshold $\beta$ governs the degree of visual similarity required for negative word selection. Smaller values of $\beta$ admit visually weak or ambiguous negatives that provide limited fine-grained supervision. In contrast, overly large $\beta$ restricts the candidate pool to a small set of near-duplicate signs, reducing training diversity and weakening the learning signal. Based on these observations, we set both $\alpha$ and $\beta$ to 0.7 in all experiments. This choice provides a balanced trade-off between alignment reliability and candidate coverage, yielding consistently strong fine-grained improvements while maintaining stable coarse-grained retrieval performance across models.

\subsection{Language-Model-Based Negative Mining Methods}

Recent studies have leveraged language models to generate hard negative captions for the vision--language retrieval task~\cite{DBLP:conf/cvpr/ranking_clip, DBLP:conf/iccv/verb_in_action, DBLP:conf/cvpr/teaching_structured_clip}. 
Following this line of work, we compare SAN with several representative language-model-based negative mining strategies, including FastText, RoBERTa, and GPT-4o-mini. For fine-grained evaluation, we fix a set of target words in advance and independently mine negative word candidates for each target word using different strategies. Each method generates exactly ten negative candidates per target word, and these candidates are subsequently used to construct method-specific fine-grained evaluation sets.

\subsubsection{FastText}
FastText~\cite{DBLP:journals/corr/JoulinGBDJM16} is a static word embedding model that represents words as compositions of character n-grams. 
We use the pretrained \textit{cc.de.300.bin} model, which is trained on German Common Crawl and Wikipedia corpora. 
To mine negative candidates, we embed all words in the dataset vocabulary and compute cosine similarities between word embeddings. 
For each target word, the top-10 most similar words are selected as negative candidates.

\subsubsection{RoBERTa}
RoBERTa~\cite{DBLP:journals/corr/roBERTa} is a transformer-based masked language model trained with dynamic masking and large-scale corpora to produce contextualized representations. 
We employ \textit{XLM-RoBERTa-base}, a multilingual variant trained on 100 languages. 
Negative candidates are obtained by replacing the target word in the original sentence with a \texttt{[MASK]} token and selecting the top-10 words with the highest unmasking probabilities.

\subsubsection{GPT-4o-mini}
GPT-4o-mini~\cite{DBLP:journals/corr/abs-2303-08774} is a lightweight autoregressive large language model designed for efficient text generation. 
We use GPT-4o-mini to generate negative word candidates by prompting the model to suggest ten distinct alternatives for each masked target word, excluding the original word. 
Unlike FastText and RoBERTa, GPT-4o-mini directly generates candidates through conditional text generation rather than embedding similarity or masked prediction.

The prompt used for negative candidate generation is shown below:

{\small
\begin{verbatim}
For each [MASK] token in the sentence, provide exactly 
10 appropriate negative candidate words.
Do NOT provide the original word or any repetitions.
Each candidate must be written in lowercase.
Do NOT include the full sentence itself. 
Output only in the following format:

You must use the origin_word exactly as provided.
<origin_word> : <candidate1>,..., <candidate10>
\end{verbatim}
}

\paragraph{Part-of-Speech Control}
Because SAN and FastText may produce negative candidates with different part-of-speech (POS) tags from the target word, we apply an additional POS filtering step during fine-grained evaluation set construction. Specifically, we use GPT-4o-mini to retain only candidates that share the same POS tag as the target word. This ensures that fine-grained evaluation focuses on visual confusability rather than trivial grammatical inconsistencies.

\subsection{Analysis of Mined Negative Words}
\begin{figure*}[p]
    \centering
    \caption[Visual examples of the mined hard negative words]{Visual examples of the mined hard negative words. Each row shows the sign gestures of the origin word, the word mined by SAN, and the word mined by other methods. The numbers in parentheses indicate visual similarity scores with respect to the origin word.}
    \label{visual_example_appendix}
    
    \begin{minipage}{0.85\textwidth}
        \centering
        \includegraphics[width=0.85\textwidth]{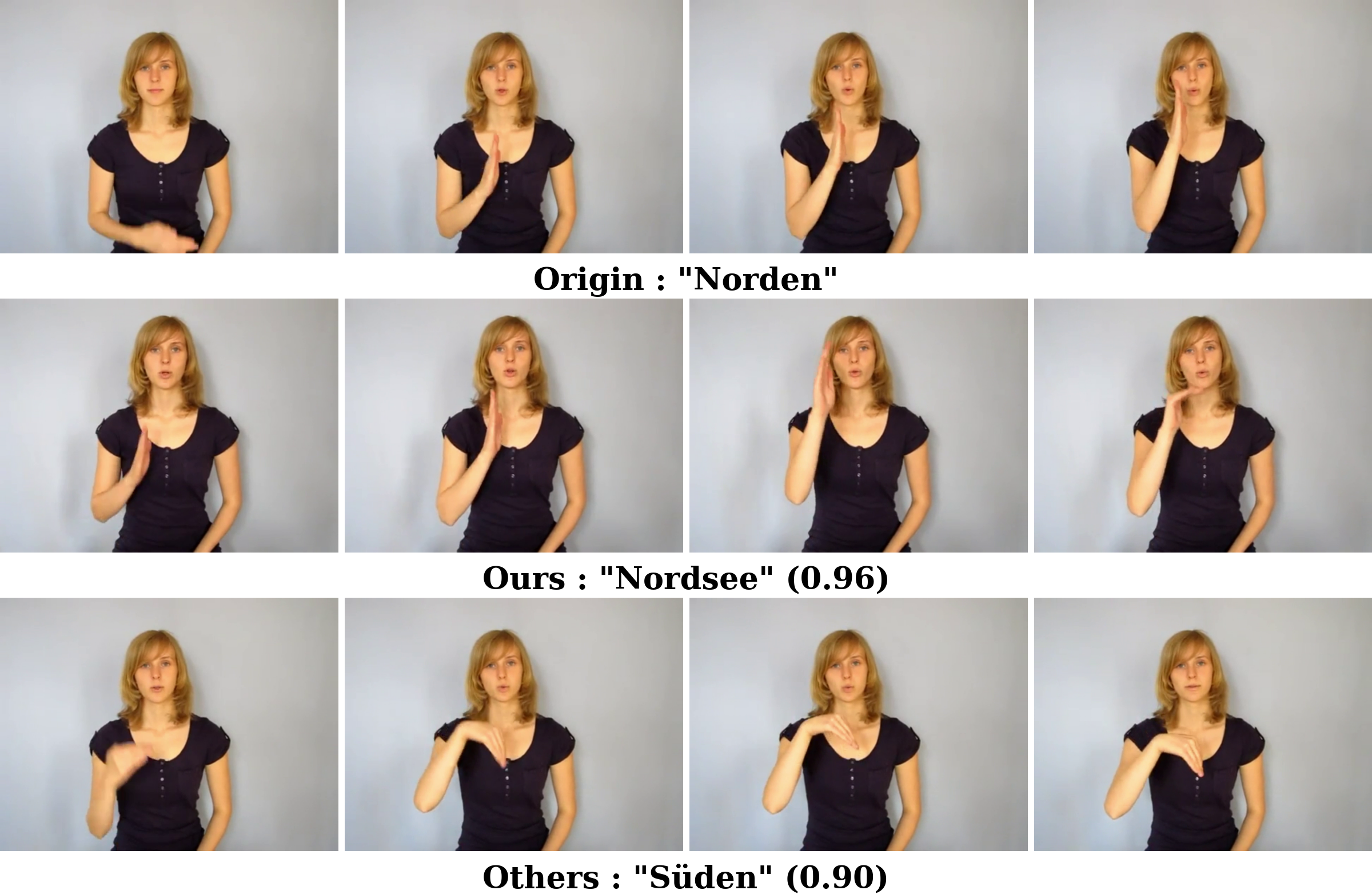}
    \end{minipage}

    \vspace{1pt}

    \begin{minipage}{0.85\textwidth}
        \centering
        \rule{0.85\textwidth}{0.4pt}\\
        \vspace{2pt}
        \includegraphics[width=0.85\textwidth]{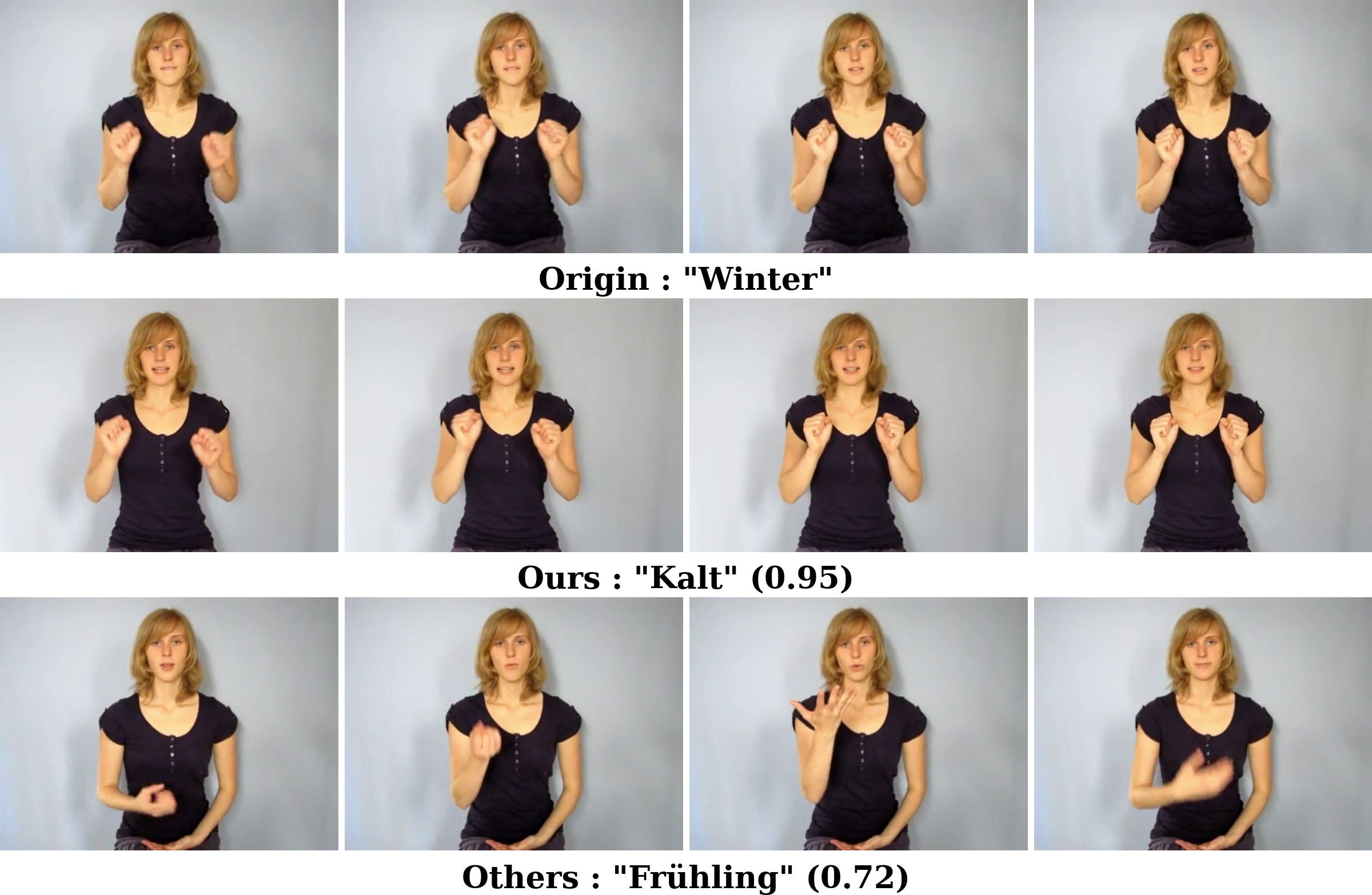}
    \end{minipage}

    \vspace{1pt}

    \begin{minipage}{0.85\textwidth}
        \centering
        \rule{0.85\textwidth}{0.4pt}\\
        \vspace{2pt}
        \includegraphics[width=0.85\textwidth]{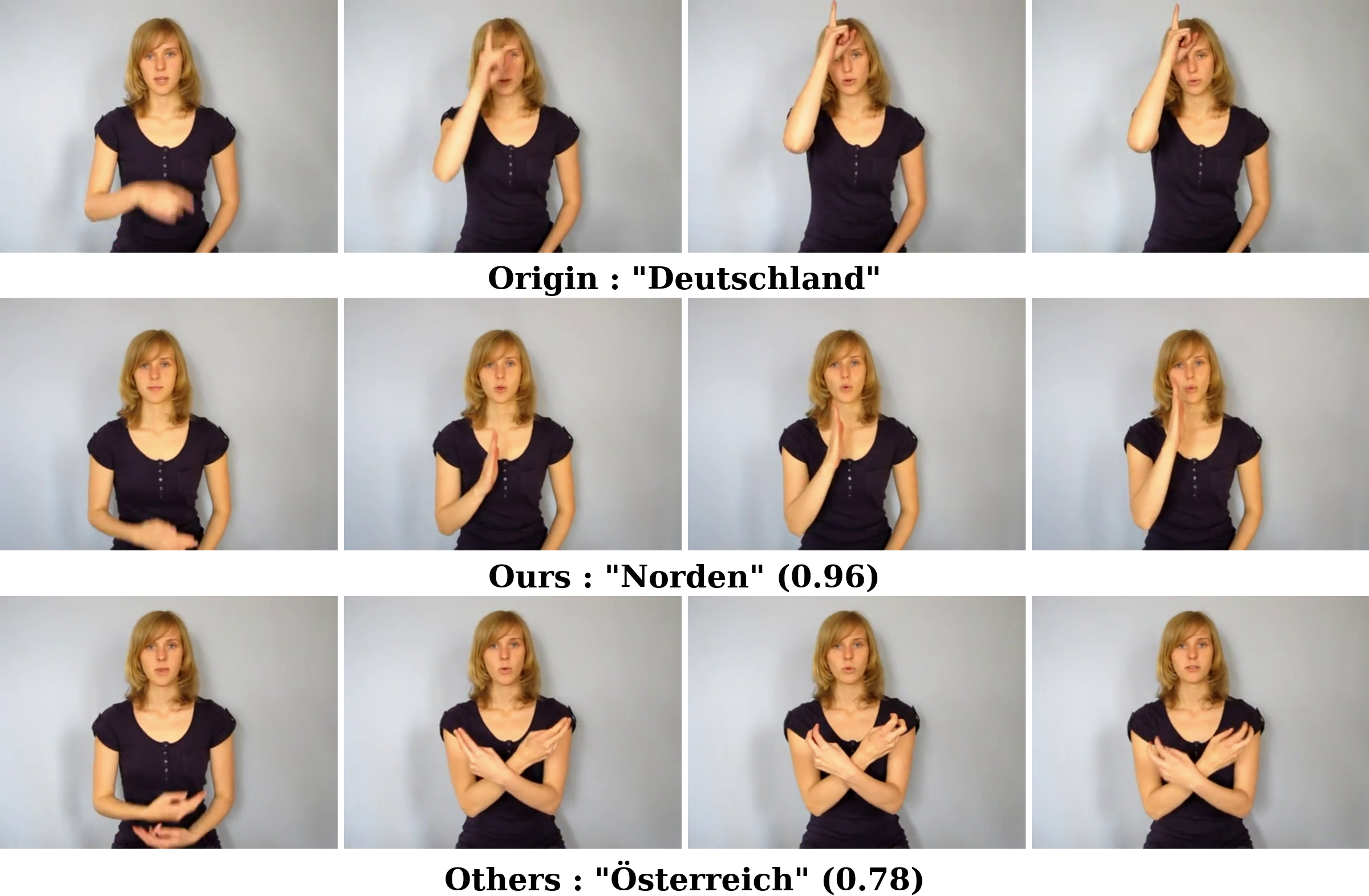}
    \end{minipage}
\end{figure*}

All sign videos in Figure~\ref{visual_example_appendix} come from the SignDict database\footnote{\url{https://signdict.org/}}.
Visual similarity scores are computed by measuring cosine similarity between sign clip features extracted by the I3D model pre-trained on the BSL-1K~\cite{DBLP:conf/cvpr/VarolMAAZ21} dataset.

For \textbf{\textit{Norden}}, SAN primarily mines words associated with northern regions, such as \textit{Nordsee}, \textit{Nordwesten}, and \textit{Nordseeküste}, whereas text-based methods tend to select direction-related words such as \textit{Osten}, \textit{Westen}, and \textit{Süden}.
In terms of signing motion, the sign for \textit{Norden} involves raising the hand vertically.
The sign for \textit{Nordsee} is visually similar, also exhibiting an upward movement, but differs subtly in the final posture, where the fingers bend by approximately $90^\circ$.
By contrast, the sign for \textit{Süden} is visually distinct from \textit{Norden}, characterized by a downward hand movement with a slightly bent wrist.

For \textbf{\textit{Winter}}, SAN mines words associated with winter-related phenomena, including \textit{kalt}, \textit{Frost}, and \textit{Wind}, while text-based approaches predominantly select season-related words such as \textit{Herbst}, \textit{Sommer}, and \textit{Frühling}.
The signs for \textit{Winter} and \textit{Kalt} are nearly identical: both involve a clenched fist, with \textit{Kalt} differing only in the absence of pursed lips and a slightly more inward-oriented fist.
In contrast, the sign for \textit{Frühling} is visually dissimilar, involving an upward unfolding motion of the right hand from a closed fist.

For \textbf{\textit{Deutschland}}, SAN mines words related to northern regions, such as \textit{Norden}, \textit{Nordsee}, and \textit{Nordmeer}, whereas text-based methods select semantically related country names including \textit{Spanien}, \textit{Europa}, and \textit{Österreich}.
Both the \textit{Deutschland} and \textit{Norden} signs involve an upward hand movement, although the hand is raised higher in the \textit{Deutschland} sign.
By contrast, the sign for \textit{Österreich} differs substantially, as it involves crossing both hands over the chest, briefly clenching the fingers, and then extending them outward.

Across these examples, negative words mined by SAN consistently exhibit higher visual similarity to the target signs than those selected by text-based methods (e.g., \textit{Nordsee} 0.96 vs.\ \textit{Süden} 0.90; \textit{Kalt} 0.95 vs.\ \textit{Frühling} 0.72; \textit{Norden} 0.96 vs.\ \textit{Österreich} 0.78).
These observations demonstrate that SAN effectively identifies hard negatives that are visually confusable with the target signs at the gesture level.

Moreover, we further compare the negative words mined by SAN and language-model-based methods in Table~\ref{tab:negative_word_example}. 
The examples show that SAN mines visually similar negatives across both semantically similar and semantically distinct word pairs. 
For example, SAN captures subtle variations among semantically related words such as \textit{Norden}--\textit{Nordsee}, as well as visually similar but semantically distinct pairs including \textit{Deutschland}--\textit{Donnerstag}, \textit{August}--\textit{Bayern}, and \textit{Europa}--\textit{Oktober}. 
These results indicate that SAN effectively mines hard negatives from visually confusable regions, covering both semantically similar and semantically distinct cases that are often missed by language-model-based methods. 
In contrast, language-model-based methods predominantly select negatives based on linguistic or semantic similarity, which may overlook visually confusable sign pairs.
For visual references of the corresponding signs, refer to the SignDict database\footnotemark.
Overall, this comparison highlights a clear difference in the types of negatives produced by each approach: SAN emphasizes visual similarity in sign gestures, while language-model-based methods focus on linguistic or semantic similarity.

\subsection{Retrieval Example}
\begin{figure}[t]
    \centering
    \begin{subfigure}[t]{0.47\linewidth}
        \centering
        \includegraphics[width=\linewidth]{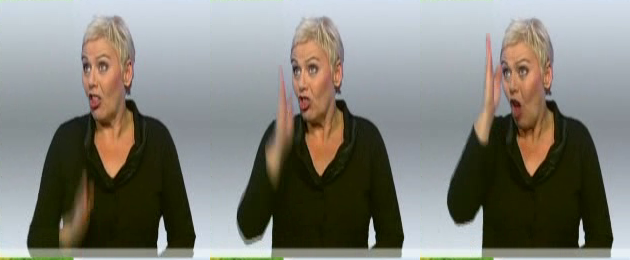}
        \caption{\textit{north}}
    \end{subfigure}
    \hfill
    \begin{subfigure}[t]{0.47\linewidth}
        \centering
        \includegraphics[width=\linewidth]{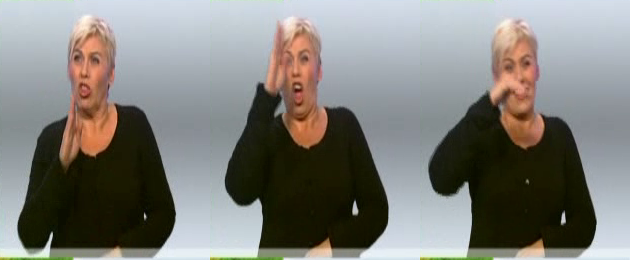}
        \caption{\textit{northwest}}
    \end{subfigure}

    \vspace{-0.5em}

    \caption{
    An example of fine-grained retrieval in PHOENIX-2014T where only SAN correctly ranks the target at R@1, while language-model-based methods fail to correctly distinguish visually confusable negatives.
    \\
    \textbf{Target caption:}
    ``In the \underline{north}, stormy gusts are also possible.''
    \\
    \textbf{Hard negative caption:}
    ``In the \underline{northwest}, stormy gusts are also possible.''
    }
    \label{fig:fine_grained_direction}
    \vspace{-0.1in}
\end{figure}

Figure~\ref{fig:fine_grained_direction} presents a representative example of fine-grained sign language retrieval on PHOENIX-2014T. 
In this example, only SAN correctly ranks the target sign video at Rank@1, while baseline methods are confused by visually similar alternatives.
The hard negative caption differs from the target caption by a single word (\textit{north} vs.\ \textit{northwest}), yet the corresponding signs exhibit high visual similarity.
This example illustrates how SAN improves fine-grained discrimination by effectively handling visually confusable sign pairs.

\clearpage
\clearpage

\clearpage 
\clearpage
\begin{table*}[p]
\centering
\caption{Comparison of mined hard negative words. SAN captures visually confusable negatives across both semantically similar and semantically distinct word pairs, whereas language-model-based methods mainly rely on linguistic similarity.}
\label{tab:negative_word_example}
\resizebox{\textwidth}{!}{%
\renewcommand{\arraystretch}{1.3}
\setlength{\tabcolsep}{15pt} 
\begin{tabular}{lll} 
\toprule
\textbf{Word} & \textbf{Ours} & \textbf{Others} \\
\midrule
Norden (North)
& Nordsee (North Sea) & Osten (East) \\
& Deutschland (Germany) & Westen (West) \\
& Nordwesten (Northwest) & Süden (South) \\
& Nordhälfte (Northern Half) & Südwesten (Southwest) \\
& Nordosthälfte (Northeastern Half) & Südosten (Southeast) \\
& Norddeutschland (Northern Germany) & Norddeutschland (Northern Germany) \\
& Nordseeküste (North Coast) & Küste (Coast) \\
\midrule        
Winter (Winter)
& Kalt (Cold) & Herbst (Autumn) \\
& Frost (Frost) & Sommer (Summer) \\
& Wind (Wind) & Frühling (Spring) \\
& Winterwetter (Winter Weather) & Schnee (Snow) \\
& Frostfrei (Frost-Free) & Jahreszeit (Season) \\
& Kälter (Colder) & Hochsommer (Midsummer) \\
& Gefrierpunkt (Freezing Point) & Frühjahr (Early Spring) \\
\midrule
Deutschland (Germany)
& Norden (North) & Spanien (Spain) \\
& Nordsee (North Sea) & Europa (Europe) \\
& Nordmeer (Northern Sea) & Österreich (Austria) \\
& Nordosten (Northeast) & England (England) \\
& Donnerstag (Thursday) & Frankreich (France) \\
& Norddeutschland (Northern Germany) & Tschechien (Czech Republic) \\
& Nordhälfte (Northern Half) & Russland (Russia) \\
\midrule                                   
August                                         
& Oktober (October) & September \\                           
& Sonntag (Sunday) & Juli (July) \\
& Bayern (Bavaria) & November \\                          
& September & Februar (February) \\ 
& Juli (July) & Juni (June) \\   
\midrule   
Europa (Europe)
& Norden (North) & Griechenland (Greece) \\
& Nordwesten (Northwest) & Schweden (Sweden) \\
& Oktober (October) & Spanien (Spain) \\
& Mitteleuropa (Central Europe) & Deutschland (Germany) \\
& S\"udosten (Southeast) & Italien (Italy) \\
& S\"udwesteuropa (Southwest Europe) & Frankreich (France) \\
& Deutschland (Germany) & Afrika (Africa) \\

\midrule                                          
Nacht (Night)
& Neun (9) & Stunde (Hour) \\
& Nordsee (North Sea) & Woche (Week) \\
& Deutschland (Germany) & Osterabend (Easter Eve) \\
& Abend (Evening) & Mitternacht (Midnight) \\     
& Donnerstag (Thursday) & Morgen (Morning) \\
\midrule
Donnerstag (Thursday)  
& Deutschland (Germany) & Montag (Monday) \\                           
& Montag (Monday)  & Freitag (Friday) \\ 
& Samstag (Saturday) & Dienstag (Tuesday) \\                 
& Norden (North) & Mittwoch (Wednesday) \\
& Sonnenschein (Sunshine) & Wochenende (Weekend) \\      


\bottomrule
\end{tabular}
}
\end{table*}
\clearpage

\end{document}